\documentclass[lettersize,journal,draftclsnofoot,onecolumn]{IEEEtran}

\usepackage{amsmath,amsfonts}
\usepackage{algorithm}
\usepackage{array}
\usepackage[caption=false,font=normalsize,labelfont=sf,textfont=sf]{subfig}
\usepackage{textcomp}
\usepackage{url}
\usepackage{verbatim}
\usepackage{graphicx}
\usepackage{cite}

\usepackage[shortlabels]{enumitem}
\usepackage{multicol,lipsum}
\usepackage{booktabs,makecell}
\usepackage[]{siunitx}
\usepackage{tikz,pgf} 
\usetikzlibrary{arrows,automata} 
\usepackage{verbatim}
\usetikzlibrary{positioning,shapes.geometric}
\usepackage{dblfloatfix}
\usepackage{algpseudocode}
\usepackage{tabularx} 
\usepackage{array}
\usepackage{float}
\usepackage{amsmath}
\usepackage{amssymb}
\usepackage{multirow}
\usepackage{mathtools,xparse}
\usepackage{amssymb}
\usepackage{array}
\usepackage{tabularx} 
\usepackage{enumitem}
\usepackage{lipsum}
\usepackage{comment}
\usepackage{color}
\usepackage{xcolor}
\usepackage{siunitx}
\usepackage{graphicx}
\usepackage{subfig}
\usepackage{amsmath}
\usepackage{multirow}
\usepackage[all,poly]{xy}
\usepackage{layouts}
\usepackage{algpseudocode}
\usepackage{array}
\usepackage{fixltx2e}
\usepackage[font=small,labelfont=bf]{caption}
\DeclarePairedDelimiter{\norm}{\lVert}{\rVert}
\newcommand{\RNum}[1]{\uppercase\expandafter{\romannumeral #1\relax}}
\def\bfDelta {{\boldsymbol{\Delta}}}

\def\bfSigma{{\boldsymbol{\Sigma}}}
\def\bfDelta{{\boldsymbol{\Delta}}}
\def\bfmu{{\boldsymbol{\mu}}}

\def\bfDelta{{\boldsymbol{\Delta}}}
\def\bfOmega{{\boldsymbol{\Omega}}}

\def\bfPsi{{\boldsymbol{\Psi}}}

\def\bs0{{\boldsymbol{0}}}

\def\bfb{{\mathbf{b}}}
\def\bfc{{\mathbf{c}}}
\def\bfd{{\mathbf{d}}}
\def\bfe{{\mathbf{e}}}

\def\bfg{{\mathbf{g}}}
\def\bfh{{\mathbf{h}}}

\def\bfu{{\mathbf{u}}}

\def\bfw{{\mathbf{w}}}
\def\bfx{{\mathbf{x}}}
\def\bfy{{\mathbf{y}}}

\def\bfA{{\mathbf{A}}}

\def\bfF{{\mathbf{F}}}

\def\bfI{{\mathbf{I}}}

\def\bfS{{\mathbf{S}}}

\def\bbC{{\mathbb{C}}}

\newcommand{\Vpixels}{\mathbf{y}}

\newcommand{\MATproj}{\boldsymbol{\Phi}}

\newcommand{\Ndistr}[1]{\mathcal{N}_{\mathbb{R}^+}\left(#1\right)}

\newcounter{algo}
\renewcommand{\thealgo}{\arabic{algo}}

\newcommand{\bSigma}{\boldsymbol{\Sigma}}

\newcommand{\bGamma}{\boldsymbol{\Gamma}}

\begin{document}


\title{Trustworthy MRI Reconstruction via Bayesian Uncertainty Quantification with Sparsity Prior Models}

\author{Ahmed Karam Eldaly,~\IEEEmembership{Member,~IEEE,} Matteo Figini, and Daniel C. Alexander

\thanks{A. K. Eldaly is with the Department of Computer Science, University of Exeter, Exeter EX4 4QF, U.K. A. K. Eldaly, M. Figini, and D. C. Alexander are with the UCL Hawkes Institute, Department of Computer Science, University College London, London, U.K.}}



\maketitle

\begin{abstract}
We propose a novel Bayesian framework for joint image reconstruction and uncertainty quantification from compressed sensing magnetic resonance imaging data. The problem is formulated as a linear inverse problem, where prior distributions are assigned to the unknown image parameters. Specifically, the image is assumed to be sparse in a given transform domain. We develop a general framework applicable to any sparsifying transform and demonstrate its performance using (1) a total variation transform based on image spatial gradients and (2) a wavelet-domain transform. Bayesian inference is performed using a split-and-augmented Gibbs sampler, while the resulting non-differentiable conditional distributions are efficiently sampled using a proximal Markov chain Monte Carlo method. The proposed algorithms are validated on both single-coil and multi-coil datasets using various k-space sampling patterns and acceleration factors. The results demonstrate that the proposed Bayesian methods consistently outperform their optimisation-based counterparts in image reconstruction while providing uncertainty estimates for the reconstructed images. Furthermore, the estimated uncertainty maps show a strong correlation with the true reconstruction errors and substantially outperformed deep learning-based uncertainty estimation methods.
\end{abstract}


\section{Introduction}
\label{sec:introduction}
\IEEEPARstart{M}{agnetic} resonance imaging (MRI) is a powerful imaging system for diagnosis of various diseases due to its high spatial resolution, noninvasive, and non-ionising radiation merits. Magnetic resonance (MR) image reconstruction from raw data involves solving the inverse problem using k-space data. There is a growing interest in developing techniques that enable faster imaging and better image quality in MRI systems. More recently, deep learning has fundamentally transformed accelerated MRI reconstruction by learning powerful image priors directly from large collections of training data, compared to classical compressed sensing image reconstruction which depends on prior assumptions regarding the underlying image \cite{lustig2007sparse, pruessmann2001advances, hong2024complex, ahmad2020plug, chen2018fast, lai2016image}. Architectures based on unrolled optimisation, variational networks, model-based deep learning, transformers, and diffusion models have consistently demonstrated superior reconstruction quality compared with traditional compressed sensing methods in terms of quantitative metrics, while simultaneously producing visually appealing reconstructions \cite{aggarwal2018modl, hammernik2018learning, tezcan2018mr, luo2020mri, korkmaz2022unsupervised, qu2024radial, cao2024hierarchical, hossain2024systematic}. However, reconstruction accuracy alone is insufficient for many clinical applications. Medical images are routinely used to support diagnosis, treatment planning, image-guided interventions, and longitudinal disease monitoring, where incorrect or over-confident reconstructions may have significant clinical consequences. Increasingly, there is growing recognition that reliable image reconstruction should not only recover the most likely image, but should also quantify the confidence associated with every reconstructed structure, thereby enabling clinicians to distinguish between reliable anatomical information and regions where the reconstruction is inherently uncertain.

The need for trustworthy and uncertainty-aware image reconstruction has recently attracted considerable attention across the medical imaging and artificial intelligence communities. Uncertainty in accelerated MRI originates from multiple sources, including aggressive k-space under-sampling, measurement noise, model mismatch, imperfect prior assumptions, and the ill-posed nature of the inverse problem itself. While several uncertainty estimation techniques have recently been proposed for deep learning, including Bayesian neural networks, Monte Carlo dropout, deep ensembles, and evidential learning, these methods primarily characterise uncertainty associated with the learning model or its parameters rather than the underlying imaging inverse problem \cite{sun2025mip, gal2016dropout, blundell2015weight, lakshminarayanan2017simple}. Furthermore, their reliability can deteriorate when the test data differ from the training distribution, a situation frequently encountered in clinical practice due to variations in scanner hardware, acquisition protocols, patient populations, and disease characteristics. Consequently, uncertainty quantification remains one of the major open challenges in the widespread adoption of AI-assisted MRI reconstruction in safety-critical clinical environments. This challenge motivates the development of principled reconstruction frameworks in which uncertainty is inferred directly from the measured k-space data and the underlying physical imaging model, providing statistically meaningful confidence measures alongside the reconstructed images.

A natural framework for uncertainty quantification in inverse problems is Bayesian inference, in which the unknown image is treated as a random variable and inference is performed by conditioning on the acquired measurements. Rather than producing a single point estimate, Bayesian methods recover the posterior distribution of the reconstructed image, enabling the computation of posterior means, credibility intervals, and spatial uncertainty maps that directly reflect the uncertainty associated with the reconstruction process. Furthermore, hierarchical Bayesian formulations naturally allow important model hyperparameters, such as regularisation parameters and noise variances, to be estimated jointly with the unknown image, avoiding the costly manual tuning typically required by optimisation-based approaches. These advantages have motivated the successful application of Bayesian inference to numerous image restoration problems, including deconvolution, super-resolution, denoising, tomography, and compressed sensing \cite{fessler2020optimization, carrillo2013sparsity, Eldaly2018Patch, eldaly2017deconvolution, eldaly2019bayesian, eldaly2021bayesian, eldaly2024Bayesian, jalal2021robust, eldaly2026bayesian}.

Despite these attractive properties, Bayesian MRI reconstruction remains considerably less mature than its optimisation-based and deep learning counterparts. The primary challenge lies in the computational complexity of posterior inference under sparsity priors. Classical compressed sensing MRI exploits sparsity in transformed domains, such as total variation and wavelet representations, whose associated prior distributions are inherently non-smooth. These priors lead to posterior distributions that are analytically intractable and notoriously difficult to sample efficiently in the high-dimensional image spaces encountered in MRI. Consequently, most existing Bayesian MRI methods introduce simplifying assumptions to render inference computationally feasible. For example, several approaches replace transform-domain sparsity with image-domain priors \cite{bilgic2011multi, chaabene2020bayesian}, while others employ approximate Bayesian inference or focus on Bayesian optimisation of k-space sampling trajectories rather than posterior uncertainty quantification of the reconstructed images \cite{serra2017parameter, haldar2019oedipus}.

This work bridges this gap by developing a general Bayesian framework for joint compressed sensing MRI reconstruction and uncertainty quantification that preserves sparsity in arbitrary transformed domains without resorting to the approximations adopted by previous Bayesian MRI methods. Thus, the main contributions can be summarised as follows. 

\begin{enumerate}
    \item We propose a Bayesian framework for joint magnetic resonance image reconstruction and uncertainty quantification. No previous work attempted to quantify the uncertainties of reconstructed images in standard compressed sensing MRI using sparsity on transformed basis.
    
    \item We formulate the magnetic resonance image reconstruction from under-sampled k-space data within a Bayesian framework and assign sparsity prior distributions in transformed domains to the unknown image field. To date, the resulting joint posterior distribution has not been explored using Markov chain Monte Carlo sampling methods.
        
    \item In addition to image reconstruction and uncertainty quantification, the proposed algorithm allows the automated estimation of the crucial model hyperparameter (regularisation parameter) which can affect the resulting reconstructed image and its uncertainty bounds.
    
    \item We demonstrate the generality of the proposed framework by considering two widely used sparsity priors: total variation (MCMC-TV) and wavelet sparsity (MCMC-Wav).

    \item We show that the proposed Bayesian framework consistently improves reconstruction quality compared with its optimisation-based maximum \textit{a posteriori} (MAP) counterparts while additionally providing statistically meaningful posterior uncertainty estimates that are unavailable from optimisation-based reconstruction methods.

    \item We demonstrate that the estimated posterior uncertainty maps exhibit strong agreement with the true reconstruction errors and provide substantially more informative uncertainty estimates than representative deep learning uncertainty quantification methods, highlighting the potential of Bayesian inference for trustworthy MRI reconstruction.
\end{enumerate}

\vspace{-0.2cm}
\section{Problem Formulation}
\label{sec:ProbForm}
\vspace{-0.1cm}
The problem of magnetic resonance image reconstruction can be expressed as follows
\begin{equation}
\begin{aligned}
\bfy = \bfA\bfx + \bfw,
\label{eq:model}
\end{aligned}    
\end{equation}
where $\bfA \in \bbC^{NL\times N}$ denotes the forward model describing a mapping from the latent image $\bfx$ to the acquired k-space data $\bfy \in \bbC^{NL}$, $\bfw \in \bbC^{NL}$ represents additive noise, and $L$ denotes the number of coils. We note that $\bfA$ consists of $L$ different submatrices $\bfA_\ell = \bfS\bfF\MATproj_\ell$ for $\ell = 1, 2, \cdots, L$, where $\bfF$ is the two-dimensional Fourier transform, $\bfS$ the k-space sampling operator, and $\MATproj_\ell$ represents the diagonal sensitivity matrix for the $\ell$-th coil. The problem investigated in this work is to estimate the image field vector $\bfx$ in Eq. \eqref{eq:model} from the observation vector $\bfy$. This inverse problem is generally ill-posed and some sort of prior information (regularisation from the optimisation perspective) is necessary to promote solutions with desired properties.

\section{Hierarchical Bayesian Model}
\label{sec:bayesian}
This section introduces the hierarchical Bayesian model developed to estimate the unknown image $\bfx$ in Eq. \eqref{eq:model}. This model is based on the likelihood function of the observations and on prior distributions assigned to the unknown parameter. The joint posterior distribution of the intensity values $\bfx$ given the observations $\bfy$ denoted as $p(\bfx|\bfy)$, can be computed using Bayes theorem as follows
\begin{equation}
\begin{aligned}
p(\bfx|\bfy) = \frac{p(\bfy | \bfx) \times p(\bfx)}{\int p(\bfy | \bfx) \times p(\bfx) d\bfx},
\end{aligned}
\end{equation}
where $p(\bfy | \bfx)$ is the likelihood of the observations, and $p(\bfx)$ is the prior distribution on $\bfx$.

\subsection{Likelihood}
The likelihood function of $\bfy$ can be expressed as
\begin{equation}
\begin{aligned}
f(\mathbf{y} \mid \mathbf{x}) = \left(\frac{1}{2 \pi \sigma^2}\right)^{\frac{NL}{2}} \exp\left( - \frac{\| \mathbf{y} - \mathbf{S} \mathbf{F} \MATproj \mathbf{x} \|_2^2 }{2 \sigma^2} \right)
\label{eq:modelG}
\end{aligned}
\end{equation}

The noise is assumed independent and identically distributed (i.i.d.) additive Gaussian noise with covariance matrix $\bSigma_{\bfy} = \sigma^2\bfI$, denoted as $\bfw \sim \mathcal{CN}(\bfw; \bf0 , \sigma^2\bfI)$, where $\bfI$ is the identity matrix.

\subsection{Parameter Prior Distributions}
The likelihood can limit the number of prior distributions adequate to result in a tractable joint posterior distribution that is easy to sample from. However, although the likelihood is quadratic, in this work, we assume that the image to be reconstructed is sparse in a given basis. The sparsity prior model can be written as follows.
\begin{equation}
\begin{aligned}
p(\bfx|\tau) = \frac{1}{\theta} \exp\left(-\tau \norm{\bfOmega(\bfx)}_1\right) \times \bf1_{\mathbb{R_+}}(\bfx),
\label{eq:L1}
\end{aligned}
\end{equation}
where $\norm{\cdot}_1$ is the $\ell_1$ norm, $\bfOmega(\bfx)$ is a transform basis applied to $\bfx$, which could be wavelets, total variation, discrete cosine transform, etc., $\theta = \int \tau \norm{\bfOmega(\bfx)}_1 d\bfx$, and $\bf1_{\mathbb{R_+}}$ is the indicator of the first orthant to impose the non-negativity constraint on the estimate. In this framework, the regularisation parameter $\tau$ in Eq. \eqref{eq:L1} is also supposed to be unknown and is included in the inference process using the stochastic approximation proximal gradient algorithm (SAPG) proposed in \cite{vidal2020maximum}.

In this work, we test two widely used sparsity bases in magnetic resonance image reconstruction: the isotropic discrete total variation (TV) regulariser and the discrete wavelet transform. The isotropic discrete TV regulariser \cite{chambolle2004algorithm} promotes piece-wise smooth images, and can be written as $\text{TV}(\bfx) = \sum_{s = 1}^{n}\sqrt{(\Delta_s^h\bfx)^2 + (\Delta_s^v\bfx)^2},$
where $\Delta_s^h\bfx$ and $\Delta_s^v\bfx$ denote the horizontal and vertical first order differences at pixel $s$, respectively. This regulariser is a discrete version of the TV regulariser proposed in \cite{rudin1992nonlinear}. Now, the prior model in Eq. \eqref{eq:L1} can be written as
\begin{equation}
\begin{aligned}
p(\bfx|\tau) = \frac{1}{\theta} \exp\left(-\tau \text{TV}(\bfx)\right) \times \bf1_{\mathbb{R_+}}(\bfx),
\label{eq:TVx}
\end{aligned}
\end{equation}
where $\theta = \int \tau \text{TV}(\bfx) d\bfx$. Alternatively, the image to be reconstructed can be assumed sparse in wavelets transform, and thus, the prior model in Eq. \eqref{eq:L1} can be written as
\begin{equation}
\begin{aligned}
p(\bfx|\tau) = \frac{1}{\theta} \exp\left(-\tau \norm{\bfPsi\bfx}_1\right) \times \bf1_{\mathbb{R_+}}(\bfx),
\label{eq:Wavx}
\end{aligned}
\end{equation}
where $\theta = \int \tau \norm{\bfPsi\bfx}_1 d\bfx$, and $\bfPsi$ denotes the wavelet transform matrix.  

\subsection{Joint Posterior Distribution}
\label{subsec:JPD}
The joint posterior of the parameter vector $\bfx$ and hyperparameter $\tau$ can be expressed as 
\begin{equation}
\begin{aligned}
p(\bfx, \tau|\Vpixels)&\propto p(\Vpixels|\bfx)p(\bfx|\tau)\propto
\left(\frac{1}{2 \pi \sigma^2}\right)^{\frac{NL}{2}} \exp\left( - \frac{\| \mathbf{y} - \mathbf{S} \mathbf{F} \MATproj \mathbf{x} \|_2^2 }{2 \sigma^2} \right)  \times \frac{1}{\theta} \exp\left(-\tau \norm{\bfOmega(\bfx)}_1\right) \times \bf1_{\mathbb{R_+}}(\bfx).
\label{eq:posteriornew}
\end{aligned}
\end{equation}


The next section describes the sampling scheme used to estimate the unknown parameter $\bfx$ and the hyperparameter $\tau$.

\section{Bayesian Inference}
\label{sec:BayesianInference}
Given the difficulty of obtaining Bayesian estimators analytically from $(f(\bfx,\tau|\bfy))$, we resort to Markov chain Monte Carlo (MCMC) sampling to draw samples from the posterior distribution in Eq. \eqref{eq:posteriornew}. In particular, we employ the split-and-augmented Gibbs sampler (SPAGS), a variable-splitting MCMC method proposed by \cite{vono2019split}, to efficiently explore the posterior distribution. For SPAGS, splitting variables are introduced to decouple the likelihood from the prior distribution, e.g. $\bfb = \bfOmega(\bfx)$, $\bfc = \bfS\bfd$, $\bfd = \bfF\bfg$, $\bfg = \MATproj\bfx$, and $\bfe = \bfx$. This splitting locally decouples each vector of variables from the remaining unknowns, thereby simplifying the inference procedure. Consequently, the joint posterior distribution in Eq. \eqref{eq:posteriornew} can be reformulated as
\begin{equation}
\begin{aligned} 
p(\bfx, \bfDelta, \bGamma| \bfy) &\propto \exp\left(-\frac{\norm{\bfy - \bfc}^2_2}{2\sigma^2}\right) \times \exp\left(-\tau \norm{\bfb}_1\right) \times \bf1_{\mathbb{R_+}}(\bfe) \\
\times & \exp\left(-\frac{\norm{\bfb - (\bfOmega(\bfx)+\bfh_{1})}_2^2}{2\rho^2}\right) \times \exp\left(-\frac{\norm{\bfh_{1}}}{2\alpha^2}\right)\\
\times & \exp\left(-\frac{\norm{\bfc - (\bfS\bfd+\bfh_{2})}_2^2}{2\rho^2}\right) \times \exp\left(-\frac{\norm{\bfh_{2}}}{2\alpha^2}\right)\\
\times & \exp\left(-\frac{\norm{\bfd - (\bfF\bfg+\bfh_{3})}_2^2}{2\rho^2}\right) \times \exp\left(-\frac{\norm{\bfh_{3}}}{2\alpha^2}\right)\\
\times & \exp\left(-\frac{\norm{\bfg - (\MATproj\bfx+\bfh_{4})}_2^2}{2\rho^2}\right) \times \exp\left(-\frac{\norm{\bfh_{4}}}{2\alpha^2}\right)\\
\times & \exp\left(-\frac{\norm{\bfe - (\bfx+\bfh_{5})}_2^2}{2\rho^2}\right) \times \exp\left(-\frac{\norm{\bfh_{5}}}{2\alpha^2}\right),
\label{eq:PosteriorSPA}
\end{aligned}
\end{equation}
where $\rho, \alpha > 0$, and $\bGamma = \{\bfh_1$, $\bfh_2, \bfh_3, \bfh_4, \bfh_5\}$ are auxiliary variables associated with $\bfDelta = \{\bfb$, $\bfc, \bfd, \bfg, \bfe\}$ respectively. We can observe that it's easy to to sample from the conditional distribution of each of the parameters. Moreover, for the non-smooth terms, we can consider the proximal Moreau-Yoshida-unadjusted Langevin algorithm (P-MYULA) proposed in \cite{durmus2018efficient}, as the conditional distribution comprises a quadratic term that is differentiable and another term that is easy to compute its proximal operator. It is worth noting here that for for the TV sparsity case, the splitting variable $\bfb = \bfx$, whereas for sparsity in wavelets, $\bfb = \bfPsi\bfx$. 

The stochastic simulation method presented in this work provides information of the full joint posterior distribution. However, in the provided experiments, we focus on two posterior statistics; the marginal posterior mean of each pixel defined as 
\begin{equation}
\begin{aligned}
E\{x_n|\bfy, \tau\} = \int_{0}^{\infty} x_n p(x_n|\bfy, \tau)d x_n,
\end{aligned}
\end{equation}
which is the minimum mean squared error estimator (MMSE) of $x_n$ \cite{robert2007bayesian}, and the marginal posterior variance defined as 
\begin{equation}
\begin{aligned}
\text{var}\{x_n|\bfy, \tau\} = \int_{0}^{\infty} \left(x_n - E\{x_n|\bfy, \tau\}\right)p(x_n|\bfy, \tau) d x_n.
\end{aligned}
\end{equation}

We propose to sample sequentially $\bfx$, $\bfDelta = \{\bfb$, $\bfc, \bfd, \bfg, \bfe\}$, $\bGamma = \{\bfh_1$, $\bfh_2, \bfh_3, \bfh_4, \bfh_5\}$, and $\tau$ using moves that are summarised in Algorithm \ref{algo:PCGS}, with details of sampling from each conditional distribution in the algorithm as follows.

\begin{algorithm}
\caption{Split-and-Augmented Gibbs Sampler for MR Joint Image Reconstruction and Uncertainty Quantification}
\label{algo:PCGS}
\begin{algorithmic}[1]
\State \textbf{Fixed input parameters}: Number of burn-in iterations $N_{\text{bi}}$, total number of iterations $N_{\text{MC}}$
\State \textbf{Initialisation} ($k = 0$)
\begin{itemize}
\item Set $\bfb^{(0)}, \bfc^{(0)}, \bfd^{(0)}, \bfg^{(0)}, \bfe^{(0)}, \bfh_1^{(0)}, \bfh_2^{(0)}, \bfh_3^{(0)}, \bfh_4^{(0)}, \bfh_5^{(0)}, \tau^{(0)}$
\end{itemize}

\State \textbf{Repeat ($1 \leq k \leq N_{\text{bi}}$)}
\item Sample $\tau^{(k)} | {\bfx}^{(k)}$ using Eq. \eqref{eq:updateTau}
\State \textbf{Set} $k = k + 1$.

\State \textbf{Repeat ($N_{\text{bi}}+1 \leq k \leq N_{\text{MC}}$)}
\item Sample $\bfx^{(k)} | \left(\bfy, \bfDelta^{(k-1)}, \bGamma^{(k-1)}, {\tau}^{(k-1)}\right)$ using Eq. \eqref{eq:SampleX}
\State Sample $\bfb^{(k)} | \left(\bfx^{(k)}, \bfDelta^{(k-1)}_{\backslash \bfb}, \bGamma^{(k-1)}\right)$ using Eq. \eqref{eq:SampleB}
\State Sample $\bfc^{(k)} | \left(\bfy, \bfx^{(k)}, \bfb^{(k)}, \bfd^{(k-1)}, \bfg^{(k-1)}, \bfe^{(k-1)}, \bGamma^{(k-1)}\right)$ using Eq. \eqref{eq:SampleC}
\State Sample $\bfd^{(k)} | \left(\bfy, \bfx^{(k)}, \bfb^{(k)}, \bfc^{(k)}, \bfg^{(k-1)}, \bfe^{(k-1)}, \bGamma^{(k-1)}\right)$ using Eq. \eqref{eq:SampleD}
\State Sample $\bfg^{(k)} | \left(\bfy, \bfx^{(k)}, \bfb^{(k)}, \bfc^{(k)}, \bfd^{(k)}, \bfe^{(k-1)}, \bGamma^{(k-1)}\right)$ using Eq. \eqref{eq:SampleG}
\State Sample $\bfe^{(k)} | \left(\bfy, \bfDelta^{(k)}_{\backslash \bfe}, \bGamma^{(k-1)}\right)$ using Eq. \eqref{eq:SampleE1}
\State Sample $\bfh_1^{(k)} | \bfDelta^{(k)}$ using Eq. \eqref{eq:SampleH1}
\State Sample $\bfh_2^{(k)} | \bfDelta^{(k)}$ using Eq. \eqref{eq:SampleHi}
\State Sample $\bfh_3^{(k)} | \bfDelta^{(k)}$ using Eq. \eqref{eq:SampleHi}
\State Sample $\bfh_4^{(k)} | \bfDelta^{(k)}$ using Eq. \eqref{eq:SampleHi}
\State Sample $\bfh_5^{(k)} | \bfDelta^{(k)}$ using Eq. \eqref{eq:SampleH5}
\State \textbf{Set} $k = k + 1$.
\end{algorithmic}
\end{algorithm}

\textit{Sampling $\bfx | \left(\bfy, \bfDelta, \bGamma, {\tau}\right)$}: It can be seen from Eq. \eqref{eq:PosteriorSPA} that the full conditional distribution of $\bfx$ reduces to 
$f\left(\bfx | \bfy, \bfDelta, \bGamma, {\tau}\right) \propto
\exp\left(-\frac{\norm{\bfb - (\bfOmega(\bfx)+\bfh_1)}_2^2}{2\rho^2}\right)
\times \exp\left(-\frac{\norm{\bfg - (\MATproj\bfx+\bfh_4)}_2^2}{2\rho^2}\right) \times \exp\left(-\frac{\norm{\bfe - (\bfx+\bfh_5)}_2^2}{2\rho^2}\right),$
which reduces to sampling from the following multivariate Gaussian distribution 
\begin{equation}
\begin{aligned}
p\left(\bfx | \bfy, \bfDelta, \bGamma, {\tau}\right) \text{     } \propto \mathcal{N}\left({\bfx; \bfmu_\bfx, \bfSigma_\bfx}\right),
\label{eq:SampleX}
\end{aligned}
\end{equation}
where
\begin{equation}
\begin{aligned}
\left\{
    \begin{array}{ll}
\boldsymbol{\mu_\bfx} = \frac{1}{\rho^2} \left( \bfOmega^T (\bfb - \bfh_1) + \MATproj^T (\bfg - \bfh_4) + (\bfe - \bfh_5) \right)\boldsymbol{\Sigma_\bfx}\\
\boldsymbol{\Sigma_\bfx} = \rho^2 \left( \bfOmega^T \bfOmega + \MATproj^T \MATproj + \bfI \right)^{-1}
\end{array}
\right.
\end{aligned}
\end{equation}
where for a TV prior model $\bfOmega = \bfI$, and for sparsity using a wavelet basis $\bfOmega = \bfPsi$. It can be noted that $\bfOmega^T\bfOmega = \bfI$, and $\MATproj^T\MATproj$ is a diagonal matrix,  therefore, it is easy to sample from Eq. \eqref{eq:SampleX} as covariance matrix is diagonal and is easy to construct.

\textit{Sampling $\bfb | \left(\bfy, \bfx, \bfDelta_{\backslash \bfb}, \bGamma\right)$}: It can be seen from Eq. \eqref{eq:PosteriorSPA} that the full conditional distribution of $\bfb$ reduces to $f\left(\bfb | \bfy, \bfx, \bfDelta_{\backslash \bfb}, \bGamma\right) \text{     }\propto  \text{     }
\frac{1}{\theta} \exp\left(-\tau \norm{\bfb}_1\right) \times \exp\left(-\frac{\norm{\bfb - (\bfOmega(\bfx)+\bfh_1)}_2^2}{2\rho^2}\right).$ Due to the non-differentiability of this conditional distribution, we consider the proximal Moreau-Yoshida-unadjusted Langevin algorithm (P-MYULA) proposed in \cite{durmus2018efficient} to generate samples asymptotically distributed according to this conditional distribution. The Markov chain of P-MYULA of a sample from the $\bfb$ variable can be written as
\begin{equation}
\begin{aligned}
\bfb^{(k+1)} &= (1 - \frac{\gamma}{\lambda})\bfb^{(k)} - \gamma \nabla f(\bfb^{(k)}) + \frac{\gamma}{\lambda}\text{prox}_g^\lambda(\bfb^{(k)})+ \sqrt{2\gamma}\,\,\Ndistr{\bf0, \bf1},
\label{eq:SampleB}
\end{aligned}
\end{equation}
\begin{equation}
\begin{aligned}
f(\bfb) = \frac{\norm{\bfb - (\bfOmega(\bfx)+\bfh_1)}_2^2}{2\rho^2} \text{, that } \nabla f(\bfb) = \frac{\bfb - \bfOmega\bfx -\bfh_1}{\rho^2},
\end{aligned}
\end{equation}
where for a TV prior model $\bfOmega = \bfI$, and for sparsity using a wavelet basis $\bfOmega = \bfPsi$. On the other hand,
\begin{equation}
\begin{aligned}
\text{prox}_g^\lambda(\bfb) = \arg \min_\bfu \tau\lambda \norm{\bfu}_1 + \frac{1}{2}\norm{\bfu - \bfb}_2^2,
\end{aligned}
\end{equation}
which in TV case, 
\begin{equation}
\begin{aligned}
\text{prox}_g^\lambda(\bfb) =  \arg \min_\bfu  \tau\lambda \text{TV}(\bfu) + \frac{1}{2}\norm{\bfu - \bfb}_2^2,
\end{aligned}
\end{equation}
which can be solved itratively using Chambolle's algorithm \cite{chambolle2004algorithm}. Whereas for sparsity in wavelets 
\begin{equation}
\begin{aligned}
\text{prox}_g^\lambda(\bfb) = \arg \min_\bfu \tau\lambda \norm{\bfu}_1 + \frac{1}{2}\norm{\bfu - \bfb}_2^2 = \text{soft}(\bfb, \tau\lambda),
\end{aligned}
\end{equation}
where $\text{soft}(\cdot, \beta)$ denotes the component-wise application of the soft-threshold function $y \rightarrow \text{sign}(y) \text{max}\{|y|-\beta, 0\}$.

\textit{Sampling {$\bfc | \left(\bfy, \bfx, \bfDelta_{\backslash \bfc}, \bGamma\right)$}:} By cancelling out the terms that do not depend on $\bfc$ in Eq. \eqref{eq:PosteriorSPA}, the full conditional distribution of $\bfc$ can be written as $p\left(\bfc | \bfy, \bfx, \bfDelta_{\backslash \bfc}, \bGamma\right) \text{     }\propto  \text{     } \exp\left(-\frac{\norm{\bfy - \bfc}^2_2}{2\sigma^2}\right) \times \exp\left(-\frac{\norm{\bfc - (\bfS\bfd+\bfh_2)}_2^2}{2\rho^2}\right),$
which reduces to sampling from the following multivariate Gaussian distribution
\begin{equation}
\begin{aligned}
p\left(\bfc | \bfy, \bfx, \bfDelta_{\backslash \bfc}, \bGamma\right) \text{     } {\propto \mathcal{CN}\left({\bfc; \bfmu_\bfc, \bfSigma_\bfc}\right),}
\label{eq:SampleC}
\end{aligned}
\end{equation}
where 
\begin{equation}
\begin{aligned}
{\bfmu_\bfc = \frac{\rho^2\bfy + \sigma^2(\bfS\bfd+\bfh_2)}{\sigma^2 + \rho^2} \text{, and    }\,\,\, \bfSigma_\bfc = \frac{\rho^2\sigma^2}{\rho^2+\sigma^2}\bfI.}
\end{aligned}
\end{equation}

\textit{Sampling {$\bfd | \left(\bfy, \bfx, \bfDelta_{\backslash \bfd}, \bGamma\right)$}:} By cancelling out the terms that do not depend on $\bfd$ in Eq. \eqref{eq:PosteriorSPA}, the full conditional distribution of $\bfd$ can be written as $p\left(\bfd | \bfy, \bfx, \bfDelta_{\backslash \bfd}, \bGamma\right) \text{     }\propto  \text{     }\exp\left(-\frac{\norm{\bfd - (\bfF\bfg+\bfh_3)}_2^2}{2\rho^2}\right) \times \exp\left(-\frac{\norm{\bfc - (\bfS\bfd+\bfh_2)}_2^2}{2\rho^2}\right),$ which reduces to sampling from the following complex multivariate Gaussian distribution
\begin{equation}
\begin{aligned}
p\left(\bfd | \bfy, \bfx, \bfDelta_{\backslash \bfd}, \bGamma\right) \text{     } {\propto \mathcal{CN}\left({\bfd; \bfmu_\bfd, \bfSigma_\bfd}\right),}
\label{eq:SampleD}
\end{aligned}
\end{equation}
where 
\begin{equation}
\begin{aligned}
\left\{
    \begin{array}{ll}
\bfmu_\bfd = \frac{1}{\rho^2}\left(\bfS^T(\bfc-\bfh_2) + \bfF(\bfg-\bfh_3)\right)\bfSigma_\bfd,\\
\bfSigma_\bfd = \rho^2\left(\bfS^T\bfS + \bfI\right)^{-1},
\end{array}
\right.
\end{aligned}
\end{equation}
which is easy to sample from as $\bfS$ is a diagonal matrix, and therefore it is easy to construct the covariance matrix $\bfSigma_\bfd$.

\textit{Sampling {$\bfg | \left(\bfy, \bfx, \bfDelta_{\backslash \bfg}, \bGamma\right)$}:} By canceling the terms that do not depend on $\bfg$ in Eq. \eqref{eq:PosteriorSPA}, the full conditional distribution of $\bfg$ can be written as $p\left(\bfg | \bfy, \bfx, \bfDelta_{\backslash \bfg}, \bGamma\right) \text{     }\propto  \text{     }\exp\left(-\frac{\norm{\bfd - (\bfF\bfg+\bfh_3)}_2^2}{2\rho^2}\right) \times \exp\left(-\frac{\norm{\bfg - (\MATproj\bfx+\bfh_4)}_2^2}{2\rho^2}\right),$ which reduces to sampling from the following complex multivariate Gaussian distribution
\begin{equation}
\begin{aligned}
p\left(\bfg | \bfy, \bfx, \bfDelta_{\backslash \bfg}, \bGamma\right) \text{     } {\propto \mathcal{CN}\left({\bfg; \bfmu_\bfg, \bfSigma_\bfg}\right),}
\label{eq:SampleG}
\end{aligned}
\end{equation}
where 
\begin{equation}
\begin{aligned}
\left\{
    \begin{array}{ll}
\boldsymbol{\mu_\bfg} = \frac{1}{\rho^2} \left( \mathbf{F}^C (\mathbf{d} - \mathbf{h}_3) + \MATproj \mathbf{x} + \mathbf{h}_4 \right)\bfSigma_\bfg,\\

\boldsymbol{\Sigma_\bfg} = \rho^2 \left( \mathbf{F}^C \mathbf{F} + \mathbf{I} \right)^{-1},
\end{array}
\right.
\end{aligned}
\end{equation}
which is easy to sample from as $\bfF^C\bfF=\bfI$, and therefore $\bfSigma_\bfg$ is a diagonal matrix that is easy to construct.

\textit{Sampling {$\bfe | \left(\bfy, \bfx, \bfDelta_{\backslash \bfe}, \bGamma\right)$}:} By cancelling out the terms that do not depend on $\bfe$ in Eq. \eqref{eq:PosteriorSPA}, the full conditional distribution of $\bfe$ can be written as
$p\left(\bfe | \bfy, \bfx, \bfDelta_{\backslash \bfe}, \bGamma\right) \propto  \exp\left(-\frac{\norm{\bfe - (\bfx+\bfh_5)}_2^2}{2\rho^2}\right) \times \bf1_{\mathbb{R_+}}(\bfe).$ In a similar fashion to sampling $\bfb$, due to the non-differentiability of this conditional distribtion, we consider P-MYULA to generate samples asymptotically distributed according to this conditional distribution. The Markov chain of P-MYULA of a new sample for $\bfe$ can be written as
\begin{equation}
\begin{aligned}
\bfe^{(k+1)} &= (1 - \frac{\gamma}{\lambda})\bfe^{(k)} - \gamma \nabla f(\bfe^{(k)}) + \frac{\gamma}{\lambda}\text{prox}_g^\lambda(\bfe^{(k)})+ \sqrt{2\gamma}\,\,\Ndistr{\bf0, \bf1},
\label{eq:SampleE1}
\end{aligned}
\end{equation}
where
\begin{equation}
\begin{aligned}
f(\bfe) = \frac{\norm{\bfe - (\bfx+\bfh_5)}_2^2}{2\rho^2}, \text{ that } \nabla f(\bfe) = \frac{\bfe - \bfx -\bfh_5}{\rho^2},
\end{aligned}
\end{equation}
and $g(\bfe) = \bf1_{\mathbb{R_+}}(\bfe),$ that 
\begin{equation}
\begin{aligned}
\text{prox}_e^\lambda(\bfe) = \arg \min_\bfu \lambda \bf1_{\mathbb{R_+}}(\bfu) + \frac{1}{2}\norm{\bfu - \bfe}_2^2 = \text{max}(\bfe, \lambda),
\end{aligned}
\end{equation}
where $\text{max}(\cdot, \beta)$ denotes the component-wise application of the max function.

\textit{Sampling {$\bfh_1 | \left(\bfx, \bfDelta\right)$}}: It can be seen that the full conditional distribution of $\bfh_1$ can be written as $p\left(\bfh_1 | \bfx, \bfDelta\right)  \propto  \exp\left(-\frac{\norm{\bfb - (\bfOmega(\bfx)+\bfh_1)}_2^2}{2\rho^2}\right) \times \exp\left(-\frac{\norm{\bfh_1}}{2\alpha^2}\right),$
which reduces to sampling from the following multivariate Gaussian distribution
\begin{equation}
\begin{aligned}
p\left(\bfh_1 | \bfx, \bfDelta\right) \text{     } {\propto \mathcal{N}\left({\bfh_1; \bfmu_{\bfh_1}, \bfSigma_{\bfh_1}}\right),}
\label{eq:SampleH1}
\end{aligned}
\end{equation}
where 
\begin{equation}
\begin{aligned}
{\bfmu_{\bfh_1} = \frac{\alpha^2}{\rho^2+\alpha^2}\left(\bfOmega \bfx - \bfb\right) \text{ and    }\,\,\, \bfSigma_{\bfh_1} = \frac{\rho^2\alpha^2}{\rho^2+\alpha^2}\bfI},
\end{aligned}
\end{equation}
where $\bfOmega = \bfI$ for a TV prior model, and $\bfOmega = \bfPsi$ for sparsity in wavelets basis.

In a similar fashion to sampling $\bfh_1$, sampling each of $\bfh_2$, $\bfh_3$ and $\bfh_4$ reduces to sampling from complex multivariate Gaussian distributions as follows.
\begin{equation}
\begin{aligned}
p\left(\bfh_i | \bfx, \bfDelta\right) \text{     } {\propto \mathcal{CN}\left({\bfh_i; \bfmu_{\bfh_i}, \bfSigma_{\bfh_i}}\right),}
\label{eq:SampleHi}
\end{aligned}
\end{equation}
with $i = \{2, 3, 4\}, $ with the following means and covariance matrices 
\begin{equation}
\begin{aligned}
\left\{
    \begin{array}{llll}
\bfmu_{\bfh_2} =  \frac{\alpha^2}{\rho^2+\alpha^2}\left( \bfS\bfd - \bfc\right),\\

\bfmu_{\bfh_3} = \frac{\alpha^2}{\rho^2+\alpha^2}\left( \bfF\bfg - \bfd\right),\\

\bfmu_{\bfh_4} = \frac{\alpha^2}{\rho^2+\alpha^2}\left( \MATproj\bfx - \bfg\right),\\

\bfSigma_{\bfh_2} = \bfSigma_{\bfh_3} = \bfSigma_{\bfh_4} = \frac{\rho^2\alpha^2}{\rho^2+\alpha^2}\bfI.
\end{array}
\right.
\end{aligned}
\end{equation}

\textit{Sampling $\bfh_5 | \left(\bfx, \bfDelta\right)$}: It can be seen that the full conditional distribution of $\bfh_5$ reduces to sampling from the following multivariate Gaussian distribution
\begin{equation}
\begin{aligned}
p\left(\bfh_5 | \bfx, \bfDelta\right) \text{     } {\propto \mathcal{N}\left({\bfh_5; \bfmu_{\bfh_5}, \bfSigma_{\bfh_5}}\right),}
\label{eq:SampleH5}
\end{aligned}
\end{equation}
where 
\begin{equation}
\begin{aligned}
\bfmu_{\bfh_5} = \frac{\alpha^2}{\rho^2+\alpha^2}\left(\bfx - \bfe\right), \text{ and } \bfSigma_{\bfh_5} = \frac{\rho^2\alpha^2}{\rho^2+\alpha^2}\bfI_N.
\end{aligned}
\end{equation}

\textit{Sampling {$\tau | \bfx$}:} Sampling the regularisation parameter $\tau|\bfx$ in either the TV or the wavelets cases is done using the stochastic approximation proximal gradient algorithm (SAPG) proposed in \cite{vidal2020maximum}. This method computes the maximum marginal likelihood using a stochastic proximal gradient optimisation algorithm that is driven by proximal MCMC samplers. The update of $\tau | \bfx$ is given as follows
\begin{equation}
\begin{aligned}
\tau^{(t+1)} = \Pi_T\left[\tau^{(t)} + \delta^{(t+1)} \left( \frac{NL}{\tau^{(t)}} - \norm{\bfOmega(\bfx^{(t)})}_1\right)\right],
\label{eq:updateTau}
\end{aligned}
\end{equation}
where $\Pi_T$ is the projection onto $T$ that is $\tau\in T$ and $\delta$ is a sequence of non-increasing step-sizes. In this work, we set $\delta$ to $\delta = \frac{0.1}{NL\times\tau^{(0)}}\times t^{-0.8}$, and $T \in [10^{-5}, 1]$ which is a standard choice in the literature \cite{vidal2020maximum, bobkov2011concentration}.


The algorithm is run for $N_{\textrm{MC}}$ iterations, of which the first $N_{\textrm{bi}}$ iterations are treated as burn-in to allow the Markov chain to reach its stationary regime. The burn-in period was determined through visual inspection of preliminary chains. Samples generated during this initial phase are discarded, while the remaining draws are retained for posterior inference. The model parameters are then estimated using the empirical mean of the retained samples, yielding minimum mean square error (MMSE) estimates. For example, the MMSE estimate of the latent intensity vector $\bfx$ is computed as

\begin{equation}
\begin{aligned}
\hat{\bfx} = \frac{1}{N_{\text{MC}} - N_{\text{bi}}}\sum_{t = N_{\text{bi} + 1}}^{N_{\text{MC}}} \bfx^{(t)}.
\end{aligned}
\end{equation}

\section{Experimental Results}
\label{sec:SyntheticData}
\subsection{Data sets}
To demonstrate the advantages of the proposed framework, we evaluate both its image reconstruction performance and its uncertainty quantification capability. Specifically, image reconstruction performance is compared with the optimisation-based counterparts of the proposed Bayesian approaches. These methods employ the same variable-splitting formulation and maximise the logarithm of the corresponding joint posterior distribution for fixed regularisation parameters, thereby producing maximum \textit{a posteriori} (MAP) estimates without quantifying the associated reconstruction uncertainty. The quality of the proposed uncertainty estimates is then evaluated against uncertainty estimation obtained from representative deep learning reconstruction networks. The evaluation is conducted using both single-coil and multi-coil MRI datasets with sub-sampled k-space measurements. For the single-coil experiments, we consider two datasets. The first is the synthetic Shepp--Logan phantom of size $256 \times 256$, which provides an exact ground-truth image and therefore enables rigorous quantitative evaluation of both reconstruction accuracy and uncertainty estimation under controlled acquisition conditions. To emulate realistic MRI acquisitions, zero-mean i.i.d. white Gaussian noise with variance $\sigma^2 = 20$ is added to the fully sampled image before generating the corresponding k-space measurements. The second dataset consists of 50 T1-weighted brain MR images from the Human Connectome Project (HCP), acquired on a 3 Tesla Siemens Connectome scanner \cite{sotiropoulos2013advances}. Each image is resized to $256 \times 256$. For the multi-coil experiments, we use the real low-field MRI dataset M4Raw \cite{lyu2023m4raw}, consisting of multi-channel brain k-space data acquired from healthy volunteers on a 0.3 Tesla MRI system. Three T2-weighted brain scans of size $256 \times 256$ are randomly selected for evaluation. The data acquisition is simulated by sub-sampling the 2D discrete Fourier transform of the three datasets, using three widely used k-space sub-sampling patterns in the literature, including 2D-random sampling, Cartesian sampling with random phase encodes (1D random), and pseudo-radial sampling. While the Cartesian and pseudo-radial sampling patterns are commonly used in actual data acquisition protocols, the 2D-random is not used in practice, but provides a useful theoretical baseline that is widely used in the literature in assessing the performance of new MR image reconstruction methodologies \cite{korkmaz2022unsupervised, zhang2020image, qu2024radial}. In our framework, the under-sampling ratios are set as \{10\%, 20\%, 30\%, 40\%, 50\%\} for Cartesian and pseudo-radial sampling, and are set as \{5\%, 10\%, 20\%, 30\%, 40\%\} for 2D-random sampling. 


\subsection{Quantitative analysis}
The two proposed approaches and the three existing methods used for comparison are run on the datasets described above. In the MCMC-TV and MCMC-Wav methods, Markov chains of length $N_{\text{bi}} = 2\times 10^4$ and a burn-in periods of length {$N_{\text{MC}} = 1.7\times 10^4$} are used. The quantitative measures used to assess the quality of reconstructed intensity fields is the root mean square error (RMSE). Different values of the hyperparameters $(\rho, \alpha)$ of the split and augmented - Gibbs sampler are tested to maximise the performance (least RMSE), and then fixed for each sub-sampling method in all experiments. The hyperparameters of the P-MYULA proximal MCMC method are set to $(\lambda, \gamma) = (\rho^2, \rho^2/4)$ as in \cite{vono2019split}. For the MCMC-Wav and ADMM-Wav methods, We use a two-level wavelet decomposition of Haar transform. Other wavelets basis are also tested but showed similar results, and therefore and not presented here. For the optimisation-based methods ADMM-TV and ADMM-Wav, several values for the regularisation parameter are tested, from which we pick the image estimate corresponding to the value providing the least RMSE. 

\subsection{Results using single-coil synthetic phantom data}
Table \ref{tab:PhantomMetrics} presents the RMSE measures for the five methods across the three under-sampling patterns. It is worth noting that the RMSE here is computed between each reconstructed image and the ground truth without the noise added. In general, each simulation-based method outperforms its counterpart optimisation-based method, except at very high under-sampling rates (10\% for the Pseudo-Radial and Cartesian sampling, and 5\% for the 2D-Random sampling). Although the ADMM-TV method in general shows promising performance at other sampling ratios, it does not allow for UQ of the estimates and also requires manual tuning of regularisation parameters, which is impractical without ground truth data. Regarding reconstruction quality using different sampling patterns at fixed sampling ratios, the Cartesian sampling pattern results in the highest RMSE values, indicating poorer quality. In contrast, the 2D-Random under-sampling pattern yields the lowest RMSE values, suggesting the best image reconstruction quality. This demonstrates that the type of under-sampling pattern and the distribution of samples significantly impact image reconstruction quality, with the 2D-Random sampling pattern providing superior results at a fixed sampling ratio.

\begin{table}
\centering
\caption{RMSE measures using the five methods, using three different under-sampling strategies, at different ratios of k-space samples using Shepp-Logan phantom. Bold indicates best results, and underlined indicate second best results.}
\label{tab:PhantomMetrics}
\begin{tabular}{cc|ccccc|}
\cline{3-7}
\multirow{3}{*}{\textbf{}} &
  \multirow{3}{*}{\textbf{}} &
  \multicolumn{5}{c|}{\textbf{Under-sampling Ratio}} \\ \cline{3-7} 
&
&
  \multicolumn{1}{c|}{\textbf{10\%}} &
  \multicolumn{1}{c|}{\textbf{20\%}} &
  \multicolumn{1}{c|}{\textbf{30\%}} &
  \multicolumn{1}{c|}{\textbf{40\%}} &
  \multicolumn{1}{c|}{\textbf{50\%}} \\ \hline\hline
\multicolumn{1}{|c|}{\multirow{5}{*}{\rotatebox[origin=c]{90}{\textbf{P-Radial}}}} &
  \textbf{IFFT} &
  \multicolumn{1}{c|}{31.15 } &
  \multicolumn{1}{c|}{22.33 } &
  \multicolumn{1}{c|}{ 17.02} &
  \multicolumn{1}{c|}{13.68 } &
  \multicolumn{1}{c|}{11.49 }  \\   \cline{2-7} 
\multicolumn{1}{|c|}{} &
  \textbf{ADMM-Wav} &
  \multicolumn{1}{c|}{{4.30}} &
  \multicolumn{1}{c|}{{2.69}} &
  \multicolumn{1}{c|}{2.44} &
  \multicolumn{1}{c|}{{2.34}} &
  \multicolumn{1}{c|}{2.31}  \\   \cline{2-7} 
\multicolumn{1}{|c|}{} &
  \textbf{ADMM-TV} &
  \multicolumn{1}{c|}{\textbf{3.48} } &
  \multicolumn{1}{c|}{\underline{2.35}} &
  \multicolumn{1}{c|}{\underline{2.21}} &
  \multicolumn{1}{c|}{\underline{2.02}} &
  \multicolumn{1}{c|}{\underline{1.92}}  \\   \cline{2-7} 
\multicolumn{1}{|c|}{} &
 \textbf{MCMC-Wav} &
  \multicolumn{1}{c|}{{4.47 }} &
  \multicolumn{1}{c|}{2.66} &
  \multicolumn{1}{c|}{{2.39}} &
  \multicolumn{1}{c|}{2.29} &
  \multicolumn{1}{c|}{{2.27}}\\   \cline{2-7} 
\multicolumn{1}{|c|}{} &
  \textbf{MCMC-TV} &
  \multicolumn{1}{c|}{\underline{3.60}} &
  \multicolumn{1}{c|}{\textbf{2.30} } &
  \multicolumn{1}{c|}{\textbf{2.14} } &
  \multicolumn{1}{c|}{\textbf{1.91} } &
  \multicolumn{1}{c|}{\textbf{1.89} } \\ \hline\hline
\multicolumn{1}{|c|}{\multirow{5}{*}{\rotatebox[origin=c]{90}{\textbf{Cartesian}}}} &
  \textbf{IFFT} &
  \multicolumn{1}{c|}{39.14 } &
  \multicolumn{1}{c|}{31.57 } &
  \multicolumn{1}{c|}{27.17 } &
  \multicolumn{1}{c|}{22.02 } &
  \multicolumn{1}{c|}{19.14 }\\   \cline{2-7} 
\multicolumn{1}{|c|}{} &
 \textbf{ADMM-Wav} &
  \multicolumn{1}{c|}{10.19 } &
  \multicolumn{1}{c|}{2.77 } &
  \multicolumn{1}{c|}{{2.58} } &
  \multicolumn{1}{c|}{{2.26} } &
  \multicolumn{1}{c|}{{2.19}} \\ 
  \cline{2-7} 
\multicolumn{1}{|c|}{} &

 \textbf{ADMM-TV} &
  \multicolumn{1}{c|}{\textbf{9.51} } &
  \multicolumn{1}{c|}{\underline{2.62}} &
  \multicolumn{1}{c|}{\underline{2.32}} &
  \multicolumn{1}{c|}{\underline{1.95}} &
  \multicolumn{1}{c|}{\underline{1.95}} \\ \cline{2-7} 
\multicolumn{1}{|c|}{} &
  \textbf{MCMC-Wav} &
  \multicolumn{1}{c|}{10.53} &
  \multicolumn{1}{c|}{3.65} &
  \multicolumn{1}{c|}{2.55} &
  \multicolumn{1}{c|}{2.20} &
  \multicolumn{1}{c|}{2.11}  \\
  \cline{2-7} 
\multicolumn{1}{|c|}{} &
  \textbf{MCMC-TV} &
  \multicolumn{1}{c|}{\underline{10.06} } &
  \multicolumn{1}{c|}{\textbf{2.53} } &
  \multicolumn{1}{c|}{\textbf{2.21}} &
  \multicolumn{1}{c|}{\textbf{1.83} } &
  \multicolumn{1}{c|}{\textbf{1.83}} \\  \hline\hline

  \multirow{3}{*}{\textbf{}} &
  \multirow{3}{*}{\textbf{}} &
  \multicolumn{5}{c|}{\textbf{Under-sampling Ratio}} \\ \cline{3-7} 
&
&
  \multicolumn{1}{c|}{\textbf{5\%}} &
  \multicolumn{1}{c|}{\textbf{10\%}} &
  \multicolumn{1}{c|}{\textbf{20\%}} &
  \multicolumn{1}{c|}{\textbf{30\%}} &
  \multicolumn{1}{c|}{\textbf{40\%}} \\ \hline\hline
\multicolumn{1}{|c|}{\multirow{5}{*}{\rotatebox[origin=c]{90}{\textbf{Random}}}} &
  \textbf{IFFT} &
  \multicolumn{1}{c|}{31.74 } &  
  \multicolumn{1}{c|}{27.85} &  
  \multicolumn{1}{c|}{23.96} &  
  \multicolumn{1}{c|}{20.36} &  
  \multicolumn{1}{c|}{17.32} \\   
  \cline{2-7} 

\multicolumn{1}{|c|}{} &
  \textbf{ADMM-Wav} &
  \multicolumn{1}{c|}{{\underline{4.93}}} &
  \multicolumn{1}{c|}{{2.81} } &
  \multicolumn{1}{c|}{{2.35}} &
  \multicolumn{1}{c|}{{2.32} } &
  \multicolumn{1}{c|}{{2.31}}  \\
  \cline{2-7} 

  \multicolumn{1}{|c|}{} &
  \textbf{ADMM-TV} &
  \multicolumn{1}{c|}{\textbf{4.89} } &
  \multicolumn{1}{c|}{\underline{2.56}} &
  \multicolumn{1}{c|}{\underline{2.16}} &
  \multicolumn{1}{c|}{\underline{2.14}} &
  \multicolumn{1}{c|}{\underline{2.14}} \\   \cline{2-7} 

\multicolumn{1}{|c|}{} &
   \textbf{MCMC-Wav} &
  \multicolumn{1}{c|}{{4.99 }} &
  \multicolumn{1}{c|}{{{2.68}}} &
  \multicolumn{1}{c|}{{2.25 }} &
  \multicolumn{1}{c|}{{2.24}} &
  \multicolumn{1}{c|}{{2.24}}  \\
  \cline{2-7} 
\multicolumn{1}{|c|}{} &
 \textbf{MCMC-TV} &
  \multicolumn{1}{c|}{\underline{4.93}} &
  \multicolumn{1}{c|}{\textbf{2.54}} &
  \multicolumn{1}{c|}{\textbf{1.82}} &
  \multicolumn{1}{c|}{\textbf{1.73} } &
  \multicolumn{1}{c|}{\textbf{1.73}}  \\ \hline
\end{tabular}
\vspace{-0.2cm}
\end{table}

\begin{figure}[ht]
    \centering
    \includegraphics[width=0.49\textwidth]{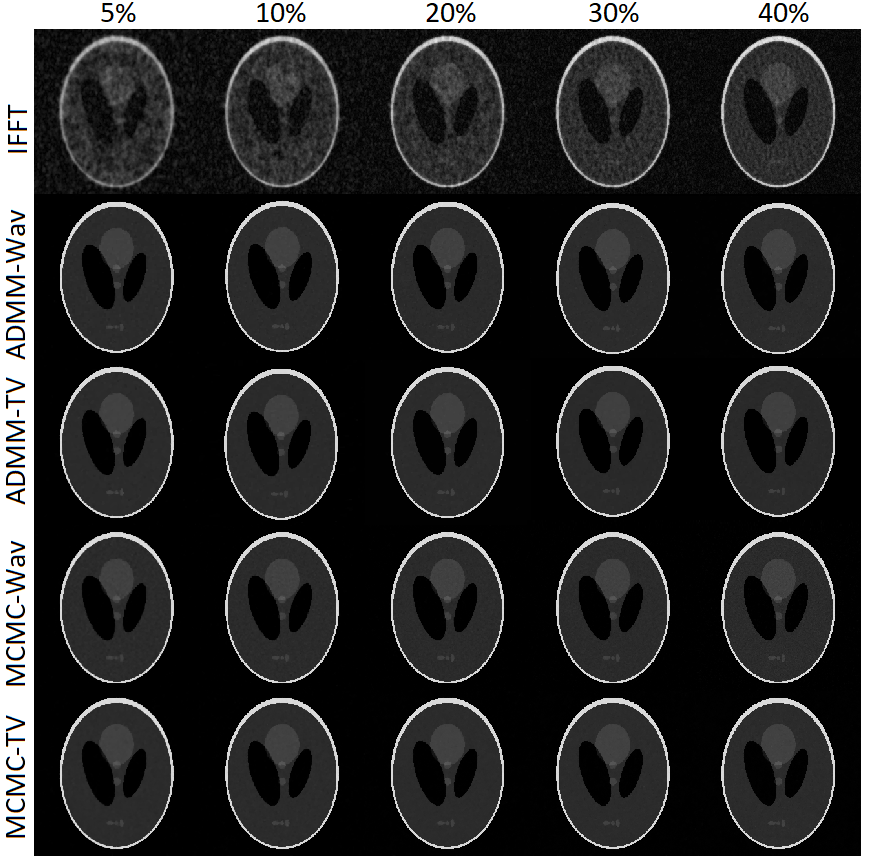}
    \vspace{-0.2cm}
    \caption{Results of reconstruction of Shepp-Logan Phantom image from using the five tested methods using the 2D-random sampling pattern with increasing under-sampling ratios.}
    \label{fig:Random_AllMethodsAllMasks_Phantom}
    \vspace{-0.4cm}
\end{figure}

\begin{figure}[ht]
    \centering
    \includegraphics[width=0.49\textwidth]{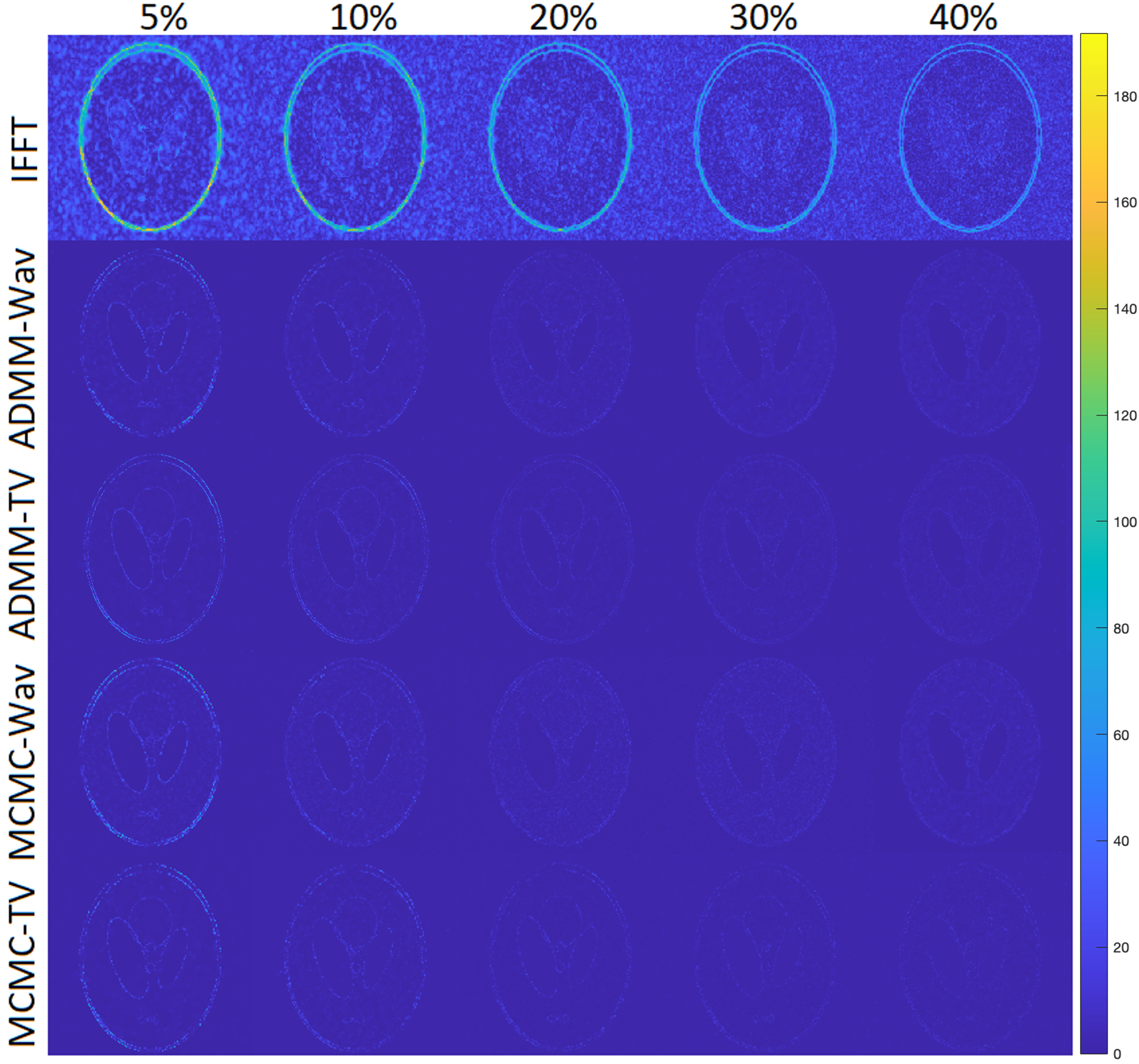}
    \vspace{-0.2cm}
    \caption{Error map images between ground truth Shepp-Logan phantom image and the image estimates in Fig. \ref{fig:Random_AllMethodsAllMasks_Phantom}.}
    \label{fig:Random_AllMethodsAllMasks_Error_Phantom}
        \vspace{-0.2cm}
\end{figure}

Figures \ref{fig:Random_AllMethodsAllMasks_Phantom} and \ref{fig:Random_AllMethodsAllMasks_Error_Phantom} show reconstruction results using the five methods, using the three under-sampling strategies at increasing under-sampling ratio, and the corresponding error map images between ground truth and each reconstructed image. The Cartesian and pseudo-radial patterns provide similar behaviour and therefore are not included in the rest of the experiments. We can observe that the baseline IFFT method shows severe artifacts due to aliasing that most of image structure is not visible. This is also clear in the error map images between the ground truth clean image and the reconstructed ones. On the other hand, the reconstruction results of the regularised methods MCMC-TV, MCMC-Wav, and their corresponding optimisation-based methods ADMM-TV and ADMM-Wav clearly devoid of the many artifacts seen in the IFFT reconstruction, despite the high under-sampling factor. Moreover, their error map images, also shows pixel errors of much smaller magnitude. Although all regularised methods could reconstruct image structure of the phantom, both the visual inspection of the reconstructed and error map images show that the MCMC-TV model better attenuates reconstruction artifacts. This is mainly clear through the error map images where the information is more structural especially next to the main reconstruction artifacts of the IFFT image. Although ADMM-TV and ADMM-Wav provide close visual results to the MCMC-TV method, they only provide point estimates, and therefore can not quantify uncertainty bounds of its estimate.

\begin{figure}[ht]
    \centering
    \includegraphics[width=0.49\textwidth]{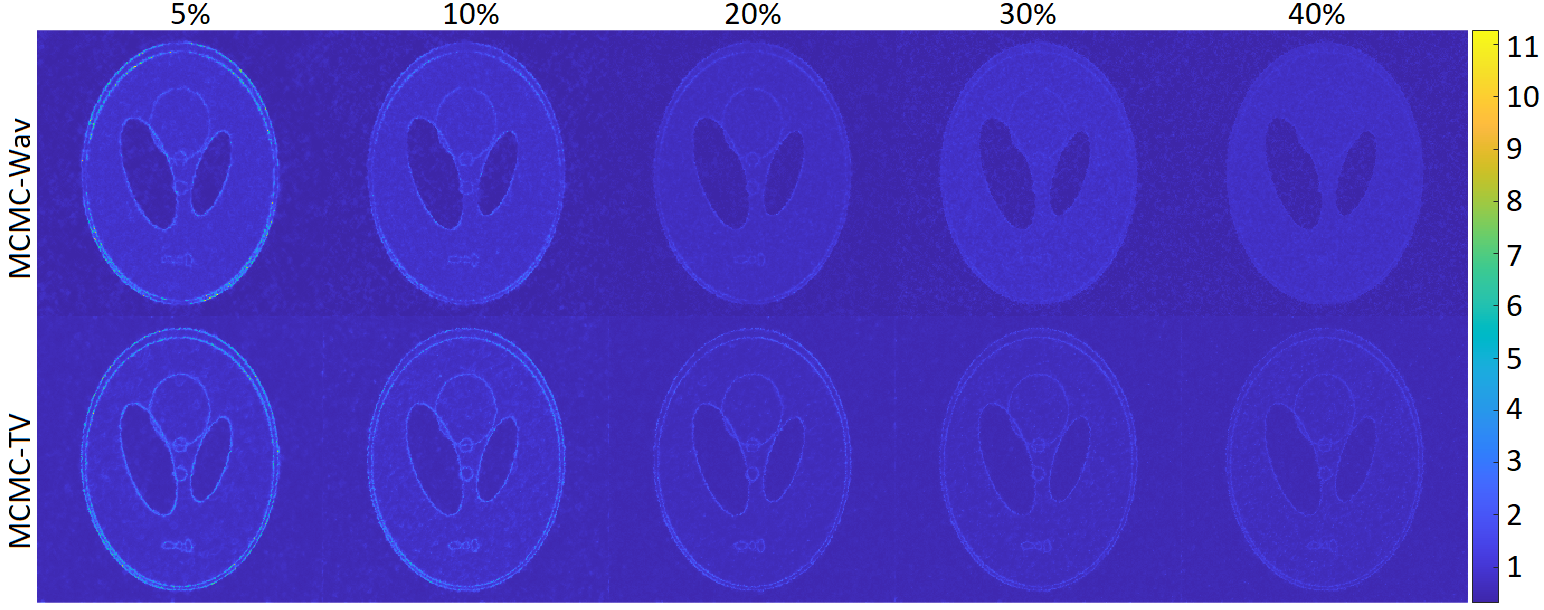}
    \vspace{-0.2cm}
    \caption{Marginal standard deviation of the MCMC-Wav and MCMC-TV estimates in Fig. \ref{fig:Random_AllMethodsAllMasks_Phantom}.}
    \label{fig:Random_AllMethodsAllMasksUQ}
    \vspace{-0.2cm}
\end{figure}

\begin{table}
\caption{Correlation coefficient computed between marginal standard deviation and error maps for the 2D-Random sampling strategy at increasing k-space sampling ratio using the Shepp-Logan phantom test image.}
\label{tab:CorrPhantom}
\centering
\begin{tabular}{l|l|l|l|l|l|}
\cline{2-6}
& \textbf{5\%} & \textbf{10\%} & \textbf{20\%} & \textbf{30\%} & \textbf{40\%} \\ \hline\hline
\multicolumn{1}{|l|}{\textbf{\begin{tabular}[c]{@{}l@{}}MCMC-Wav\end{tabular}}} & 0.742     & 0.727  &  0.706  &   0.700   &   0.691  \\ \hline\hline
\multicolumn{1}{|l|}{\textbf{\begin{tabular}[c]{@{}l@{}}MCMC-TV\end{tabular}}} & 0.754 &  0.748  &  0.740  &  0.711  & 0.707  \\ \hline
\end{tabular}
\vspace{-0.2cm}
\end{table}

As discussed earlier, the simulation-based methods MCMC-TV and MCMC-Wav draw samples that are asymptotically distributed according to their joint posterior distribution. These samples can be utilised to quantify the uncertainty bounds of their posterior mean. Figure \ref{fig:Random_AllMethodsAllMasksUQ} displays the marginal standard deviation of both methods at increasing under-sampling ratios. It is evident that as the sampling ratio increases, the uncertainty of the estimates also increases, and vice versa. The uncertainty maps correlate well with the error maps shown in Figure \ref{fig:Random_AllMethodsAllMasks_Error_Phantom} for the MCMC-TV and MCMC-Wav methods. Specifically, as the k-space under-sampling ratio increases, the errors become more pronounced, and the uncertainty of the estimates rises correspondingly. This correlation is quantified in Table \ref{tab:CorrPhantom}, which presents the correlation coefficients between the error maps and the marginal standard deviations. The results reveal that the MCMC-TV method exhibits a higher correlation between the error maps and marginal standard deviation compared to the MCMC-Wav method. This finding aligns with the earlier RMSE results, where the MCMC-TV method demonstrated superior quantitative performance over MCMC-Wav. The notable correlation with the calculated errors supports the hypothesis that the estimated uncertainty effectively captures the true uncertainty. Consequently, this uncertainty metric can serve as a reliable tool to aid in image interpretation, especially when no ground truth is available.

\subsection{Results using single-coil real brain MR images from the HCP dataset}
We applied the five reconstruction methods to reconstruct 50 real brain MRI data from the HCP dataset using three k-space under-sampling patterns at increasing under-sampling ratios. The RMSE results, summarised in Table \ref{tab:HCPMetrics}, follow a similar trend to those observed with the Shepp-Logan phantom. In particular, the IFFT method consistently results in the highest RMSE, indicating the poorest reconstruction quality. In contrast, the MCMC-TV method achieves the best reconstruction results, with the lowest RMSE values across all tested under-sampling patterns and k-space ratios, except at very high under-sampling rates (10\% for Pseudo-radial and Cartesian sampling and 5\% for 2D-random sampling). Consistent with the Shepp-Logan phantom results, the Cartesian sampling pattern generally produces the highest RMSE, indicating lower reconstruction quality, whereas the 2D-random under-sampling pattern yields the lowest RMSE, suggesting the best image reconstruction quality. This underscores the significant impact of the under-sampling pattern and sample distribution on reconstruction performance.

\begin{figure}
    \centering
\includegraphics[width=0.49\textwidth]{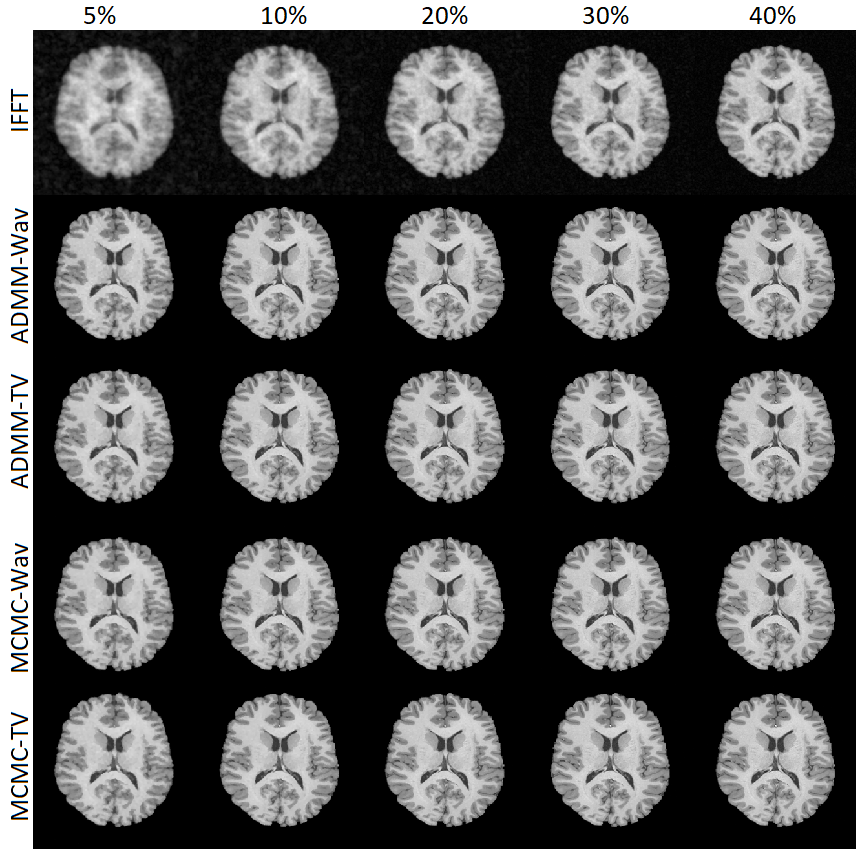}
\vspace{-0.15cm}
\caption{Results of reconstruction of a brain image from HCP data set using the five tested methods using the 2D-random sampling pattern with increasing under-sampling ratios.}
\label{fig:Random_AllMethodsAllMasksHCP}
\vspace{-0.4cm}
\end{figure}
\begin{figure}
\centering
\includegraphics[width=0.49\textwidth]{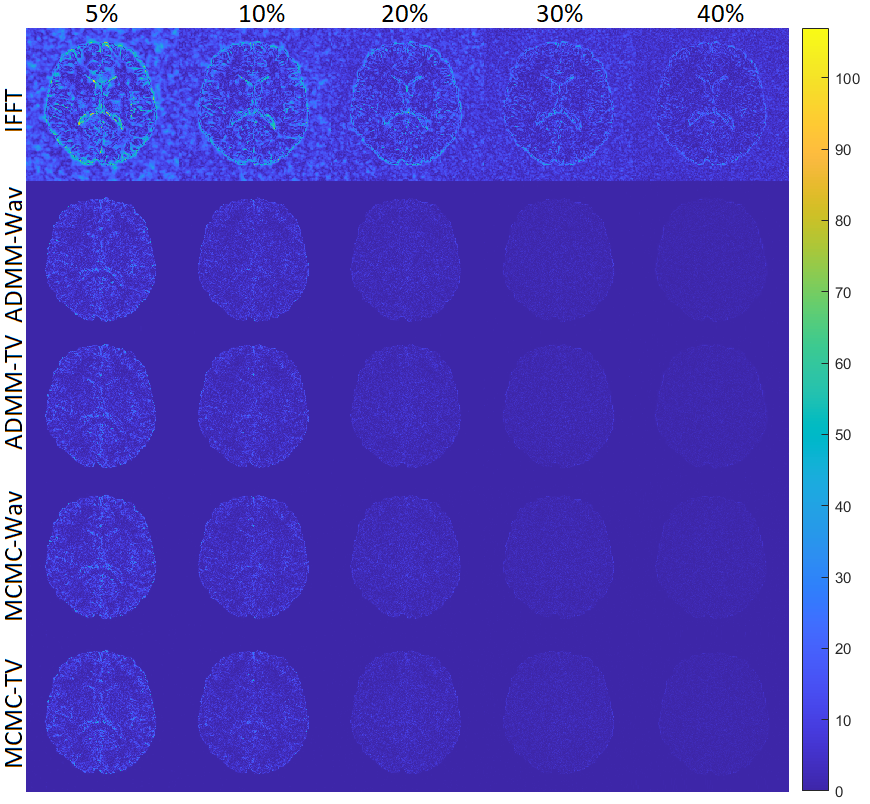}
\vspace{-0.15cm}
\caption{Error map images between ground truth brain image and the image estimates in Fig. \ref{fig:Random_AllMethodsAllMasksHCP}.}
\label{fig:Random_AllMethodsAllMasks_Errors_HCP}
\vspace{-0.4cm}
\end{figure}
\begin{figure}
    \centering
    \includegraphics[width=0.49\textwidth]{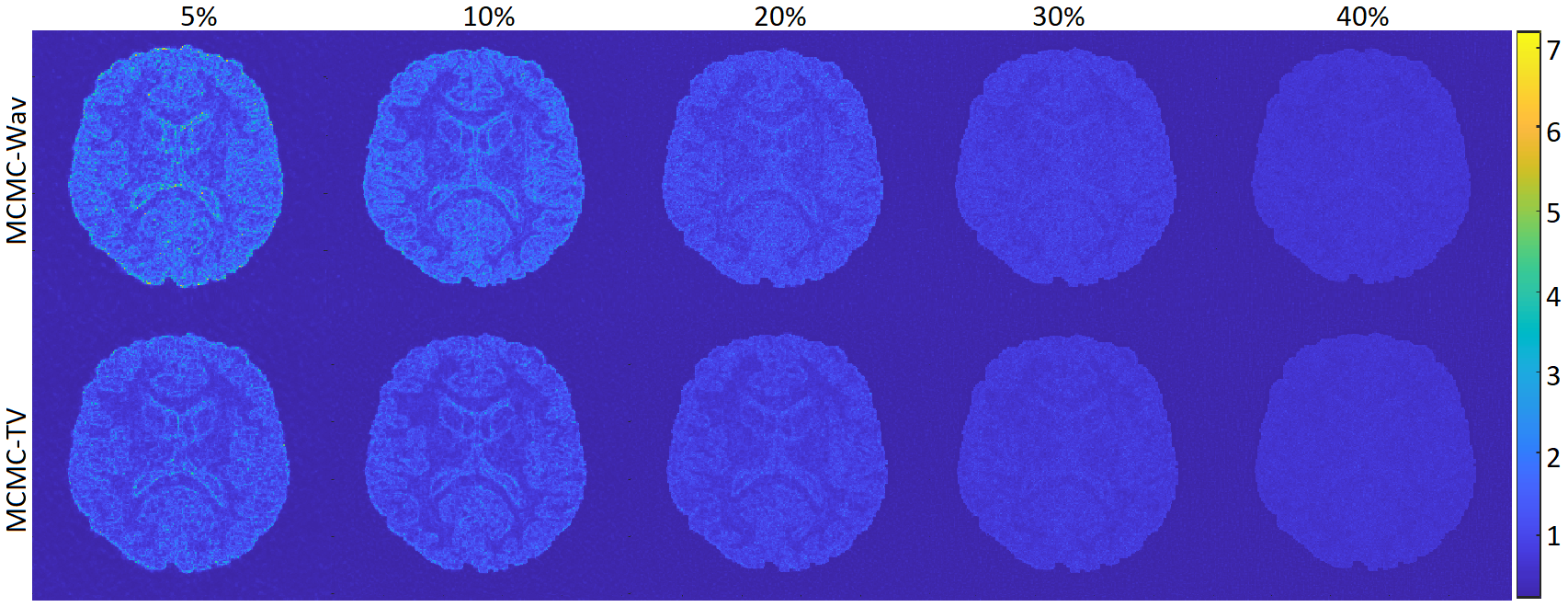}
    \vspace{-0.15cm}
    \caption{Marginal standard deviation of the MCMC-Wav and MCMC-TV estimates in Fig. \ref{fig:Random_AllMethodsAllMasksHCP} at increasing under-sampling ratio.}
    \label{fig:Random_AllMethodsAllMasksUQ_HCP}
    \vspace{-0.4cm}
\end{figure}

\begin{table}
\centering
\caption{Average RMSE measures, and standard deviation (shown in brackets) using the five methods, using three different under-sampling strategies, at different ratios of k-space samples using HCP brain images. Bold indicates best results, and underlined indicate second best results.}
\label{tab:HCPMetrics}
\begin{tabular}{cc|ccccc|}
\cline{3-7}
\multirow{3}{*}{} &
  \multirow{3}{*}{} &
  \multicolumn{5}{c|}{\textbf{Under-sampling Ratio}} \\ \cline{3-7} 
 &
   &
  \multicolumn{1}{c|}{\textbf{10\%}} &
  \multicolumn{1}{c|}{\textbf{20\%}} &
  \multicolumn{1}{c|}{\textbf{30\%}} &
  \multicolumn{1}{c|}{\textbf{40\%}} &
  \multicolumn{1}{c|}{\textbf{50\%}} \\ \hline\hline
\multicolumn{1}{|c|}{\multirow{10}{*}{\rotatebox[origin=c]{90}{\textbf{P-Radial}}}} &
  \textbf{IFFT} &
  \multicolumn{1}{c|}{\begin{tabular}[c]{@{}c@{}}16.73\\(2.76)\end{tabular}} &
\multicolumn{1}{c|}{\begin{tabular}[c]{@{}c@{}}10.23\\(2.20)\end{tabular}} &
\multicolumn{1}{c|}{\begin{tabular}[c]{@{}c@{}}7.15\\(1.90)\end{tabular}} &
\multicolumn{1}{c|}{\begin{tabular}[c]{@{}c@{}}5.35\\(1.50)\end{tabular}} &
\multicolumn{1}{c|}{\begin{tabular}[c]{@{}c@{}}4.05\\(1.10)\end{tabular}} \\ \cline{2-7}
\multicolumn{1}{|c|}{} &
\textbf{\begin{tabular}[c]{@{}l@{}}ADMM\\ -Wav\end{tabular}} &
\multicolumn{1}{c|}{\begin{tabular}[c]{@{}c@{}}7.75\\(1.70)\end{tabular}} &
\multicolumn{1}{c|}{\begin{tabular}[c]{@{}c@{}}3.95\\(1.40)\end{tabular}} &
\multicolumn{1}{c|}{\begin{tabular}[c]{@{}c@{}}2.96\\(1.20)\end{tabular}} &
\multicolumn{1}{c|}{\begin{tabular}[c]{@{}c@{}}2.17\\(0.90)\end{tabular}} &
\multicolumn{1}{c|}{\begin{tabular}[c]{@{}c@{}}1.78\\(0.70)\end{tabular}} \\ \cline{2-7}
\multicolumn{1}{|c|}{} &
\textbf{\begin{tabular}[c]{@{}l@{}}ADMM\\ -TV\end{tabular}} &
\multicolumn{1}{c|}{\textbf{\begin{tabular}[c]{@{}c@{}}7.45\\(1.60)\end{tabular}}} &
\multicolumn{1}{c|}{\underline{\begin{tabular}[c]{@{}c@{}}3.82\\(1.30)\end{tabular}}} &
\multicolumn{1}{c|}{\underline{\begin{tabular}[c]{@{}c@{}}2.70\\(1.10)\end{tabular}}} &
\multicolumn{1}{c|}{\underline{\begin{tabular}[c]{@{}c@{}}1.96\\(0.85)\end{tabular}}} &
\multicolumn{1}{c|}{\underline{\begin{tabular}[c]{@{}c@{}}1.50\\(0.60)\end{tabular}}} \\ \cline{2-7}
\multicolumn{1}{|c|}{} &
\textbf{\begin{tabular}[c]{@{}l@{}}MCMC\\ -Wav\end{tabular}} &
\multicolumn{1}{c|}{\begin{tabular}[c]{@{}c@{}}7.87\\(1.70)\end{tabular}} &
\multicolumn{1}{c|}{\begin{tabular}[c]{@{}c@{}}4.42\\(1.50)\end{tabular}} &
\multicolumn{1}{c|}{\begin{tabular}[c]{@{}c@{}}2.95\\(1.20)\end{tabular}} &
\multicolumn{1}{c|}{\begin{tabular}[c]{@{}c@{}}2.15\\(0.90)\end{tabular}} &
\multicolumn{1}{c|}{\begin{tabular}[c]{@{}c@{}}1.76\\(0.70)\end{tabular}} \\ \cline{2-7}
\multicolumn{1}{|c|}{} &
\textbf{\begin{tabular}[c]{@{}l@{}}MCMC\\ -TV\end{tabular}} &
\multicolumn{1}{c|}{\underline{\begin{tabular}[c]{@{}c@{}}7.57\\(1.65)\end{tabular}}} &
\multicolumn{1}{c|}{\textbf{\begin{tabular}[c]{@{}c@{}}3.80\\(1.25)\end{tabular}}} &
\multicolumn{1}{c|}{\textbf{\begin{tabular}[c]{@{}c@{}}2.68\\(1.05)\end{tabular}}} &
\multicolumn{1}{c|}{\textbf{\begin{tabular}[c]{@{}c@{}}(1.91)\\(0.80)\end{tabular}}} &
\multicolumn{1}{c|}{\textbf{\begin{tabular}[c]{@{}c@{}}1.41\\(0.55)\end{tabular}}} \\ \hline\hline
\multicolumn{1}{|c|}{\multirow{10}{*}{\rotatebox[origin=c]{90}{\textbf{Cartesian}}}} &
\textbf{IFFT} &
\multicolumn{1}{c|}{\begin{tabular}[c]{@{}c@{}}24.70\\(3.20)\end{tabular}} &
\multicolumn{1}{c|}{\begin{tabular}[c]{@{}c@{}}16.64\\(2.80)\end{tabular}} &
\multicolumn{1}{c|}{\begin{tabular}[c]{@{}c@{}}12.19\\(2.40)\end{tabular}} &
\multicolumn{1}{c|}{\begin{tabular}[c]{@{}c@{}}9.38\\(2.00)\end{tabular}} &
\multicolumn{1}{c|}{\begin{tabular}[c]{@{}c@{}}7.50\\(1.60)\end{tabular}} \\ \cline{2-7}
\multicolumn{1}{|c|}{} &
\textbf{\begin{tabular}[c]{@{}l@{}}ADMM\\ -Wav\end{tabular}} &
\multicolumn{1}{c|}{\begin{tabular}[c]{@{}c@{}}13.70\\(2.50)\end{tabular}} &
\multicolumn{1}{c|}{\begin{tabular}[c]{@{}c@{}}6.62\\(2.00)\end{tabular}} &
\multicolumn{1}{c|}{\begin{tabular}[c]{@{}c@{}}5.27\\(1.60)\end{tabular}} &
\multicolumn{1}{c|}{\begin{tabular}[c]{@{}c@{}}2.64\\(1.20)\end{tabular}} &
\multicolumn{1}{c|}{\begin{tabular}[c]{@{}c@{}}1.42\\0.80\end{tabular}}\\ \cline{2-7}
\multicolumn{1}{|c|}{} &

\textbf{\begin{tabular}[c]{@{}l@{}}ADMM\\ -TV\end{tabular}} & 
\multicolumn{1}{c|}{\textbf{\begin{tabular}[c]{@{}c@{}}13.08\\(2.10)\end{tabular}}} & 
\multicolumn{1}{c|}{\underline{\begin{tabular}[c]{@{}c@{}}6.50\\(1.80)\end{tabular}}} & 
\multicolumn{1}{c|}{\underline{\begin{tabular}[c]{@{}c@{}}4.54\\(1.60)\end{tabular}}} & 
\multicolumn{1}{c|}{\underline{\begin{tabular}[c]{@{}c@{}}2.16\\(1.20)\end{tabular}}} & 
\multicolumn{1}{c|}{\underline{\begin{tabular}[c]{@{}c@{}}1.35\\(0.90)\end{tabular}}} \\ \cline{2-7} 
\multicolumn{1}{|c|}{} & 
\textbf{\begin{tabular}[c]{@{}l@{}}MCMC\\ -Wav\end{tabular}} & 
\multicolumn{1}{c|}{\begin{tabular}[c]{@{}c@{}}13.83\\(2.10)\end{tabular}} & 
\multicolumn{1}{c|}{\begin{tabular}[c]{@{}c@{}}6.59\\(1.90)\end{tabular}} & 
\multicolumn{1}{c|}{\begin{tabular}[c]{@{}c@{}}5.25\\(1.70)\end{tabular}} & 
\multicolumn{1}{c|}{\begin{tabular}[c]{@{}c@{}}2.60\\(1.30)\end{tabular}} & 
\multicolumn{1}{c|}{\begin{tabular}[c]{@{}c@{}}1.38\\(0.95)\end{tabular}} \\ \cline{2-7} 
\multicolumn{1}{|c|}{} & 
\textbf{\begin{tabular}[c]{@{}l@{}}MCMC\\ -TV\end{tabular}} & 
\multicolumn{1}{c|}{\underline{\begin{tabular}[c]{@{}c@{}}13.18\\(2.00)\end{tabular}}} & 
\multicolumn{1}{c|}{\textbf{\begin{tabular}[c]{@{}c@{}}6.48\\(1.80)\end{tabular}}} & 
\multicolumn{1}{c|}{\textbf{\begin{tabular}[c]{@{}c@{}}4.51\\(1.50)\end{tabular}}} & 
\multicolumn{1}{c|}{\textbf{\begin{tabular}[c]{@{}c@{}}2.13\\(1.20)\end{tabular}}} & 
\multicolumn{1}{c|}{\textbf{\begin{tabular}[c]{@{}c@{}}1.31\\(0.85)\end{tabular}}} \\ \hline\hline
\multicolumn{1}{l}{} &
  \multicolumn{1}{l|}{} &
  \multicolumn{5}{c|}{\textbf{Under-sampling Ratio}} \\ \cline{3-7}
\multicolumn{1}{l}{} &
  \multicolumn{1}{l|}{} &
  \multicolumn{1}{c|}{\textbf{5\%}} &
  \multicolumn{1}{c|}{\textbf{10\%}} &
  \multicolumn{1}{c|}{\textbf{20\%}} &
  \multicolumn{1}{c|}{\textbf{30\%}} &
  \multicolumn{1}{c|}{\textbf{40\%}}  \\ \hline\hline

\multicolumn{1}{|c|}{\multirow{10}{*}{\rotatebox[origin=c]{90}{\textbf{Random}}}} &
\textbf{IFFT} & 
\multicolumn{1}{c|}{\begin{tabular}[c]{@{}c@{}}19.85\\(2.80)\end{tabular}} & 
\multicolumn{1}{c|}{\begin{tabular}[c]{@{}c@{}}15.32\\(2.50)\end{tabular}} & 
\multicolumn{1}{c|}{\begin{tabular}[c]{@{}c@{}}12.08\\(2.20)\end{tabular}} & 
\multicolumn{1}{c|}{\begin{tabular}[c]{@{}c@{}}9.72\\(1.80)\end{tabular}} & 
\multicolumn{1}{c|}{\begin{tabular}[c]{@{}c@{}}7.67\\(1.50)\end{tabular}} \\ \cline{2-7} 
\multicolumn{1}{|c|}{} & 
\textbf{\begin{tabular}[c]{@{}l@{}}ADMM\\ -Wav\end{tabular}} & 
\multicolumn{1}{c|}{\begin{tabular}[c]{@{}c@{}}5.62\\(1.50)\end{tabular}} & 
\multicolumn{1}{c|}{\begin{tabular}[c]{@{}c@{}}3.71\\(1.20)\end{tabular}} & 
\multicolumn{1}{c|}{\begin{tabular}[c]{@{}c@{}}2.26\\(1.00)\end{tabular}} & 
\multicolumn{1}{c|}{\begin{tabular}[c]{@{}c@{}}1.45\\(0.80)\end{tabular}} & 
\multicolumn{1}{c|}{\underline{\begin{tabular}[c]{@{}c@{}}0.98\\(0.60)\end{tabular}}} \\ \cline{2-7} 
\multicolumn{1}{|c|}{} & 
\textbf{\begin{tabular}[c]{@{}l@{}}ADMM\\ -TV\end{tabular}} & 
\multicolumn{1}{c|}{\textbf{\begin{tabular}[c]{@{}c@{}}5.56\\(1.50)\end{tabular}}} & 
\multicolumn{1}{c|}{\underline{\begin{tabular}[c]{@{}c@{}}3.68\\(1.20)\end{tabular}}} & 
\multicolumn{1}{c|}{\underline{\begin{tabular}[c]{@{}c@{}}2.25\\(1.00)\end{tabular}}} & 
\multicolumn{1}{c|}{\underline{\begin{tabular}[c]{@{}c@{}}1.42\\(0.80)\end{tabular}}} & 
\multicolumn{1}{c|}{\underline{\begin{tabular}[c]{@{}c@{}}0.98\\(0.60)\end{tabular}}} \\ \cline{2-7} 
\multicolumn{1}{|c|}{} & 
\textbf{\begin{tabular}[c]{@{}l@{}}MCMC\\ -Wav\end{tabular}} & 
\multicolumn{1}{c|}{\begin{tabular}[c]{@{}c@{}}5.70\\(1.60)\end{tabular}} & 
\multicolumn{1}{c|}{\begin{tabular}[c]{@{}c@{}}3.70\\(1.20)\end{tabular}} & 
\multicolumn{1}{c|}{\underline{\begin{tabular}[c]{@{}c@{}}2.25\\(1.00)\end{tabular}}} & 
\multicolumn{1}{c|}{\begin{tabular}[c]{@{}c@{}}1.44\\(0.80)\end{tabular}} & 
\multicolumn{1}{c|}{\underline{\begin{tabular}[c]{@{}c@{}}0.98\\(0.60)\end{tabular}}} \\ \cline{2-7} 
\multicolumn{1}{|c|}{} & 
\textbf{\begin{tabular}[c]{@{}l@{}}MCMC\\ -TV\end{tabular}} & 
\multicolumn{1}{c|}{\underline{\begin{tabular}[c]{@{}c@{}}5.65\\(1.60)\end{tabular}}} & 
\multicolumn{1}{c|}{\textbf{\begin{tabular}[c]{@{}c@{}}3.63\\(1.20)\end{tabular}}} & 
\multicolumn{1}{c|}{\textbf{\begin{tabular}[c]{@{}c@{}}2.17\\(1.00)\end{tabular}}} & 
\multicolumn{1}{c|}{\textbf{\begin{tabular}[c]{@{}c@{}}1.33\\(0.80)\end{tabular}}} & 
\multicolumn{1}{c|}{\textbf{\begin{tabular}[c]{@{}c@{}}0.96\\(0.60)\end{tabular}}} \\ \hline  
\end{tabular}
\vspace{-0.1cm}
\end{table}

\begin{table}
\caption{Average correlation coefficient, with standard deviation shown in second row, computed between marginal standard deviation and error maps for the 2D-Random sampling strategy at increasing k-space sampling ratio using brain test images from the HCP data set.}
\vspace{-0.2cm}
\label{tab:CorrHCP}
\centering
\begin{tabular}{c|c|c|c|c|c|}
\cline{2-6}
& \textbf{5\%} & \textbf{10\%} & \textbf{20\%} & \textbf{30\%} & \textbf{40\%} \\ \hline\hline
\multicolumn{1}{|l|}{\textbf{\begin{tabular}[c]{@{}l@{}}MCMC-Wav\end{tabular}}} 
& \begin{tabular}[c]{@{}c@{}}0.735\\(0.064)\end{tabular} & 
\begin{tabular}[c]{@{}c@{}}0.721\\(0.060)\end{tabular} & 
\begin{tabular}[c]{@{}c@{}}0.717\\(0.052)\end{tabular} & 
\begin{tabular}[c]{@{}c@{}}0.705\\(0.050)\end{tabular} & 
\begin{tabular}[c]{@{}c@{}}0.700\\(0.048)\end{tabular} \\ \hline\hline
\multicolumn{1}{|l|}{\textbf{\begin{tabular}[c]{@{}l@{}}MCMC-TV\end{tabular}}}
& \begin{tabular}[c]{@{}c@{}}0.748\\(0.068)\end{tabular} & 
\begin{tabular}[c]{@{}c@{}}0.740\\(0.059)\end{tabular} & 
\begin{tabular}[c]{@{}c@{}}0.730\\(0.051)\end{tabular} & 
\begin{tabular}[c]{@{}c@{}}0.710\\(0.047)\end{tabular} & 
\begin{tabular}[c]{@{}c@{}}0.702\\(0.045)\end{tabular} \\ \hline
\end{tabular}
\vspace{-0.4cm}
\end{table}

Figures \ref{fig:Random_AllMethodsAllMasksHCP} and \ref{fig:Random_AllMethodsAllMasks_Errors_HCP} show reconstruction results of a brain image from the HCP using the five approaches and the three under-sampling strategies at increasing under-sampling ratios, and error map images between the original clean image and the reconstructed ones, respectively. The other tested images show similar behaviours and therefore are not included here. For the MR image with rich texture, the IFFT method consistently exhibits prominent streaking artifacts. In contrast, the regularised methods (MCMC-TV, MCMC-Wav, ADMM-TV, and ADMM-Wav) produce reconstructions with significantly fewer artifacts. Among these, the MCMC-TV approach delivers the most satisfactory results (except at very high sampling ratio), characterised by clear contours, sharp edges, and fine image details. This is particularly evident in the error map images, where the structural information is more coherent, especially near the major artifacts present in the IFFT reconstructions. The MCMC-TV method effectively reduces these artifacts, resulting in more accurate and reliable reconstructions. These qualitative observations align with the earlier quantitative results, where MCMC-TV demonstrated superior performance in terms of lower RMSE values. 

Figure \ref{fig:Random_AllMethodsAllMasksUQ_HCP} shows the marginal standard deviation of the MCMC-TV and MCMC-Wav methods with increasing under-sampling ratios. The results confirm the trends observed with the synthetic Shepp-Logan phantom image. Specifically, as the sampling ratio increases, the uncertainty of the estimates also increases, and vice versa. The uncertainty maps correlate well with the error maps shown in Figure \ref{fig:Random_AllMethodsAllMasks_Errors_HCP} for both the MCMC-TV and MCMC-Wav methods. As the k-space under-sampling ratio increases, the errors become more pronounced, and the uncertainty of the estimates rises correspondingly. This relationship is quantified in Table \ref{tab:CorrHCP}, which shows the correlation coefficients between the error map images and the marginal standard deviations. The MCMC-TV method exhibits a higher correlation between the error maps and marginal standard deviation compared to the MCMC-Wav method. This finding aligns with the quantitative results discussed earlier, where MCMC-TV demonstrated superior quantitative performance over MCMC-Wav. The notable correlation with the calculated errors supports the hypothesis that the estimated uncertainty effectively captures the true uncertainty, making it a reliable metric for aiding image interpretation in practice when no ground truth is available.

\subsection{Results using multi-coil using real low-field MRI data from the M4Raw dataset}
\begin{figure}
\centering
\includegraphics[width=0.44\textwidth]{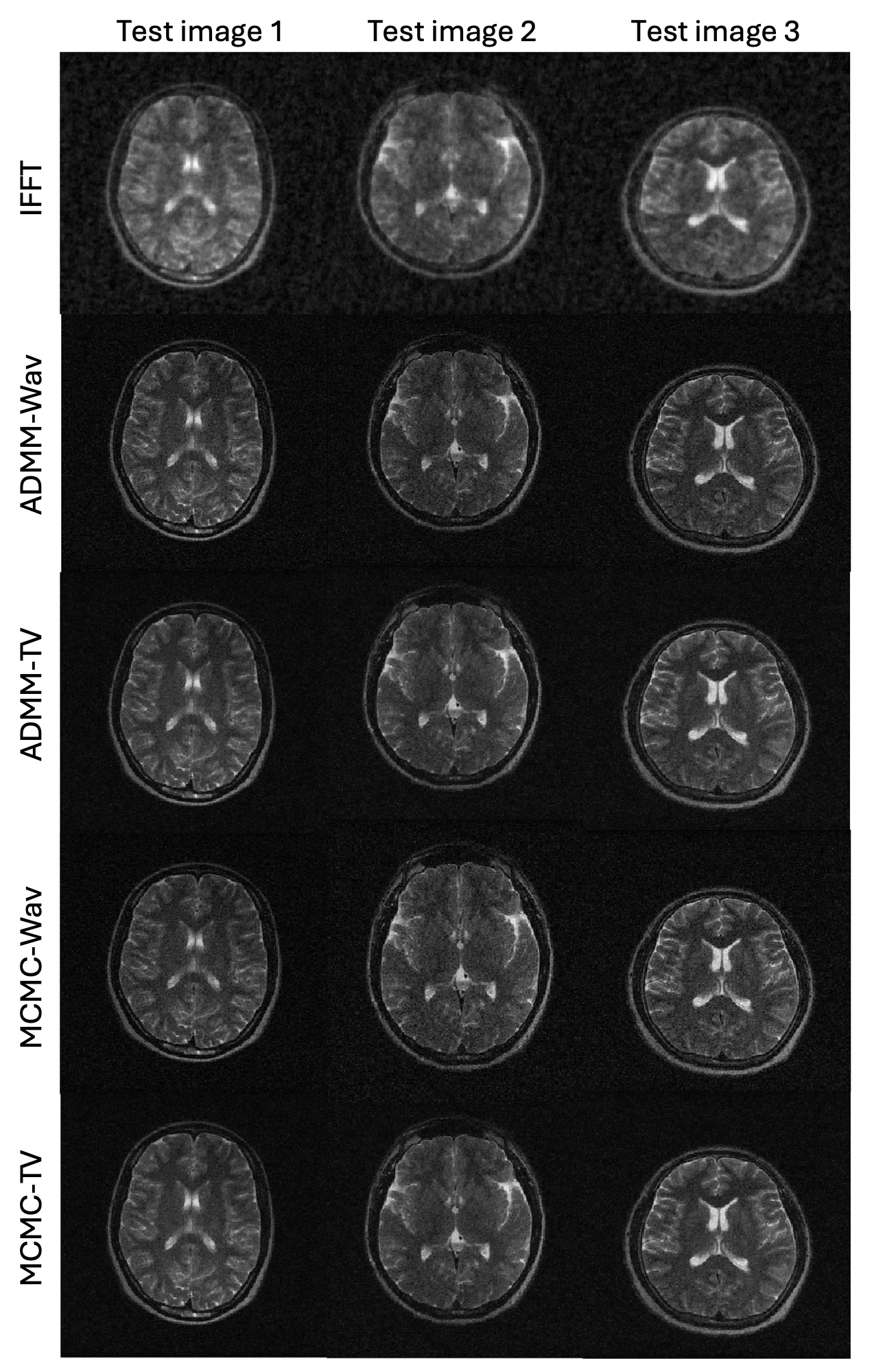} \caption{Results of reconstruction of three brain test images from the real low-field MRI M4RAW data set using the five tested methods using the 2D-random sampling pattern at 20\% under-sampling ratio.}
\label{fig:RealLowField}
\vspace{-0.4cm}
\end{figure}

\begin{figure}
    \centering
\includegraphics[width=0.44\textwidth]{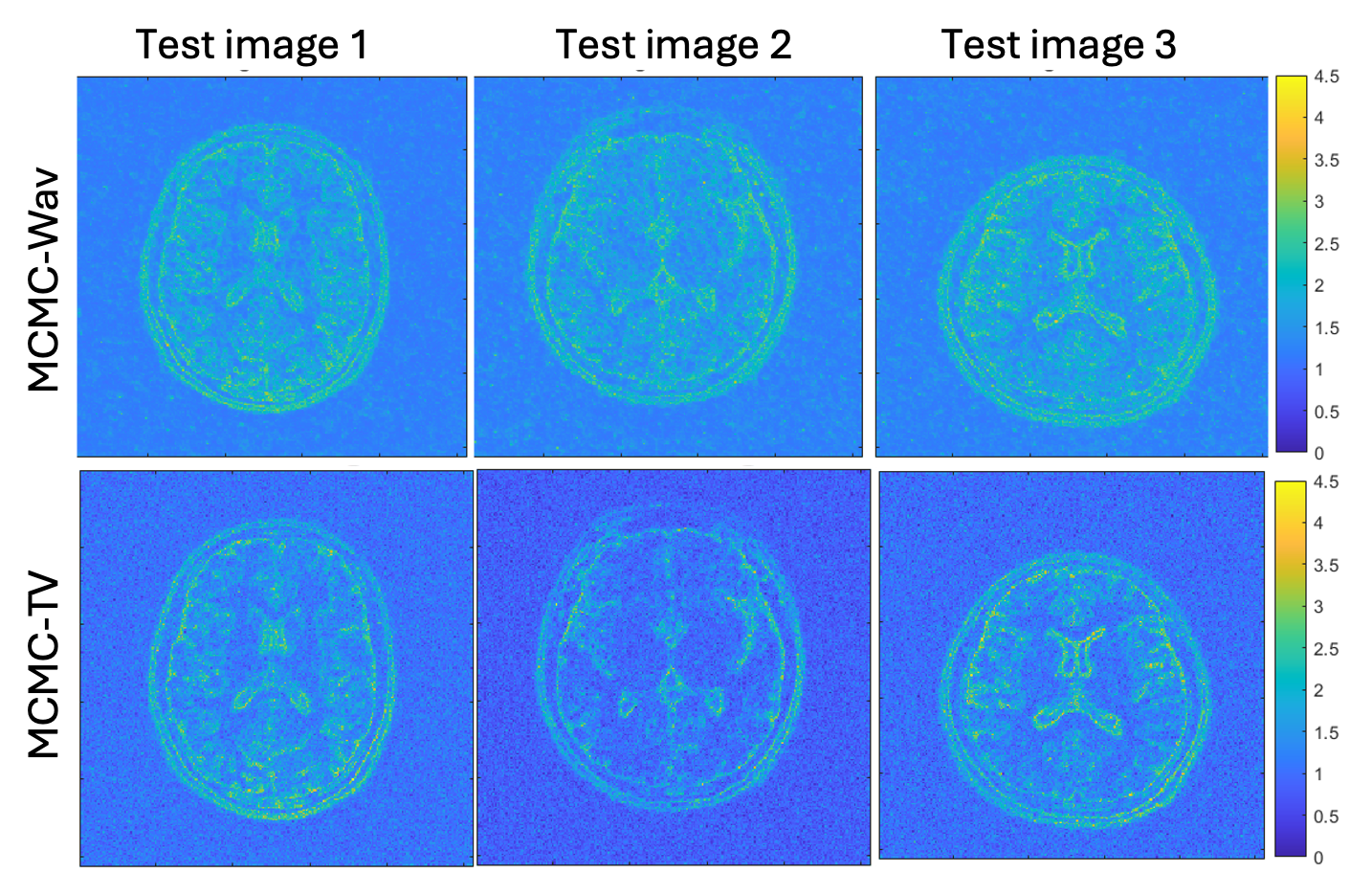} \caption{Marginal standard deviation of the MCMC-Wav and MCMC-TV estimates in Fig. \ref{fig:RealLowField}.}
\label{fig:RealLowFieldUQ}
\vspace{-0.4cm}
\end{figure}

The proposed framework is generic and can be used for image reconstruction from multiple coils. We applied the proposed approaches and the four baseline methods to reconstruct real low-field MRI k-space data from the M4Raw data set. As in the experiments with the synthetic phantom and the HCP data set, we test the three sub-sampling patterns; 2D-random, Cartesian, and pseudo-radial at increasing sampling rates. Due to similar performance across patterns and the increasing sampling rate as to the previous experiments, we report results for the 2D-random sub-sampling pattern at a 20\% sub-sampling rate. Since no ground truth is available, we present image reconstruction results alongside the corresponding uncertainty maps for the Bayesian methods. Figure \ref{fig:RealLowField} displays the reconstructed images for the selected k-space data using these methods. While the optimisation-based methods yield competitive reconstructions to the Bayesian methods, they only provide point estimates and lack uncertainty quantification. In contrast, the Bayesian methods offers uncertainty bounds, as shown in Fig. \ref{fig:RealLowFieldUQ}, which illustrates the marginal standard deviation. The results indicate higher uncertainty near intensity transitions, consistent with expected behavior, while regions of constant intensity exhibit lower uncertainty.

\subsection{Comparison with deep learning-based uncertainty quantification methods}
Although deep learning methods currently achieve state-of-the-art image reconstruction performance, the objective of this study is not to compare reconstruction accuracy alone, but rather to evaluate the reliability of the estimated uncertainty. Specifically, we investigate whether the estimated uncertainty accurately reflects the true reconstruction error, which is essential for trustworthy MRI reconstruction. To this end, we compare the proposed Bayesian framework with two representative deep learning reconstruction networks, MoDL \cite{aggarwal2018modl} and VarNet \cite{hammernik2018learning}, where predictive uncertainty is estimated using Deep Ensembles (DEns) \cite{lakshminarayanan2017simple}. Deep Ensembles estimate predictive uncertainty by independently training multiple instances of the same network using different random initialisations and combining their predictions. The variability across the ensemble predictions is then used as a measure of predictive uncertainty.

The comparison was performed using the HCP dataset, as it provides a sufficiently large number of images for training the deep learning models while also providing ground-truth images that enable quantitative assessment of uncertainty quality through comparison with reconstruction error maps. Such a comparison was not possible using the synthetic Shepp-Logan phantom dataset, which consists of a single image, nor with the M4Raw dataset, where fully sampled ground-truth images are unavailable. Table \ref{tab:ComaprisonDL} compares the correlation coefficients between the estimated uncertainty maps and the corresponding reconstruction error maps for the HCP brain dataset. The results demonstrate a clear advantage for the proposed Bayesian frameworks. Both MCMC-TV and MCMC-Wav consistently achieve high correlations across all sampling ratios. In contrast, the Deep Ensemble implementations of MoDL and VarNet achieve substantially lower correlations. The gap between the compressed sensing and deep learning approaches remains large across all sampling ratios, with the Bayesian methods typically exhibiting approximately two to three times higher correlation with the true reconstruction error. One possible explanation is that the uncertainty estimated by the proposed framework is directly linked to the inverse problem formulation and the posterior distribution of the reconstructed image given the measured k-space data. Consequently, the uncertainty primarily reflects ambiguity arising from the acquisition process, including under-sampling and measurement noise. In contrast, Deep Ensemble uncertainty is derived from variations between independently trained neural networks and therefore additionally captures uncertainty associated with model parameter estimation, network weights, optimisation, and training data characteristics. While such uncertainties are relevant for assessing model behaviour, the results suggest that they are less strongly associated with the actual reconstruction error in the recovered images. These findings support the hypothesis stated in \ref{sec:introduction} that uncertainty quantification based on Bayesian compressed sensing can provide a more reliable characterisation of reconstruction uncertainty than uncertainty estimates derived from deep learning models. Therefore the proposed methods provide greater practical value when interpreting reconstructed images in the absence of ground-truth data.

\begin{table}
\caption{Comparison of uncertainty quality between the proposed Bayesian compressed sensing methods and deep learning-based methods using deep ensembles (DEns) uncertainty estimation.}
\label{tab:ComaprisonDL}
\centering
\begin{tabular}{c|c|c|c|c|c|}
\cline{2-6}
& \textbf{5\%} & \textbf{10\%} & \textbf{20\%} & \textbf{30\%} & \textbf{40\%} \\ \hline\hline

\multicolumn{1}{|l|}{\textbf{MCMC-Wav}}
& \begin{tabular}[c]{@{}c@{}}0.735\\(0.064)\end{tabular} & 
\begin{tabular}[c]{@{}c@{}}0.721\\(0.060)\end{tabular} & 
\begin{tabular}[c]{@{}c@{}}0.717\\(0.052)\end{tabular} & 
\begin{tabular}[c]{@{}c@{}}0.705\\(0.050)\end{tabular} & 
\begin{tabular}[c]{@{}c@{}}0.700\\(0.048)\end{tabular} \\ \hline
\multicolumn{1}{|l|}{\textbf{\begin{tabular}[c]{@{}l@{}}MCMC-TV\end{tabular}}}
& \begin{tabular}[c]{@{}c@{}}0.748\\(0.068)\end{tabular} & 
\begin{tabular}[c]{@{}c@{}}0.740\\(0.059)\end{tabular} & 
\begin{tabular}[c]{@{}c@{}}0.730\\(0.051)\end{tabular} & 
\begin{tabular}[c]{@{}c@{}}0.710\\(0.047)\end{tabular} & 
\begin{tabular}[c]{@{}c@{}}0.702\\(0.045)\end{tabular}
\\ \hline\hline

\multicolumn{1}{|l|}{\textbf{MoDL (DEns)}}
& \begin{tabular}[c]{@{}c@{}}0.397\\(0.017)\end{tabular}
& \begin{tabular}[c]{@{}c@{}}0.388\\(0.012)\end{tabular}
& \begin{tabular}[c]{@{}c@{}}0.327\\(0.017)\end{tabular}
& \begin{tabular}[c]{@{}c@{}}0.269\\(0.021)\end{tabular}
& \begin{tabular}[c]{@{}c@{}}0.202\\(0.015)\end{tabular}
\\ \hline

\multicolumn{1}{|l|}{\textbf{VarNet (DEns)}}
& \begin{tabular}[c]{@{}c@{}}0.336\\(0.027)\end{tabular}
& \begin{tabular}[c]{@{}c@{}}0.303\\(0.014)\end{tabular}
& \begin{tabular}[c]{@{}c@{}}0.278\\(0.016)\end{tabular}
& \begin{tabular}[c]{@{}c@{}}0.243\\(0.027)\end{tabular}
& \begin{tabular}[c]{@{}c@{}}0.236\\(0.031)\end{tabular}
\\ \hline
\end{tabular}
\vspace{-0.2cm}
\end{table}

\vspace{-0.2cm}
\subsection{Limitations}
Despite the advantages of the proposed approach, a few limitations remain. First, the computational cost is higher than that of optimisation-based reconstruction methods because posterior inference requires generating a large number of MCMC samples. Nevertheless, this additional computational cost enables automatic estimation of the regularisation parameter and principled posterior uncertainty quantification that are unavailable in optimisation-based methods. For instance, the average computation time of the proposed MCMC-based framework ranges from approximately 7 to 9.5 minutes per image, whereas the optimisation-based methods require only 45–55 seconds per reconstruction (MATLAB implementation on a laptop equipped with a 2.8 GHz CPU, 16 GB RAM, running Microsoft Windows 10). Moreover, the reported execution time for the optimisation-based methods corresponds to a single optimisation run. In practice, multiple reconstructions with different regularisation parameter values are typically required to identify an appropriate parameter, whereas the proposed Bayesian framework estimates this parameter automatically during inference. Second, although modern deep learning reconstruction methods generally achieve superior image quality when sufficient training data are available, the objective of the proposed framework is not to maximise reconstruction accuracy alone, but to provide statistically meaningful uncertainty estimates alongside the reconstructed image. The estimated uncertainty arises directly from the imaging forward model, the measurement noise, and the assumed image prior. In contrast, uncertainty estimates obtained from deep learning are additionally influenced by uncertainty in the learned model parameters and the training data distribution. Finally, the proposed framework does not require supervised training data, making it particularly attractive in applications where annotated datasets are limited or difficult to acquire. 

\vspace{-0.2cm}
\section{Conclusions}
\label{sec:Conclusion}
This paper presented a Bayesian framework for joint magnetic resonance image reconstruction and uncertainty quantification from under-sampled k-space data. The problem is formulated within a Bayesian framework, and a sparse prior model was assigned to the unknown image field in transformed domain. Bayesian inference was performed using a split and augmented - Gibbs sampler, with the non-smooth terms were sampled using a proximal Markov chain Monte Carlo method. The proposed Bayesian methods consistently outperformed their optimisation-based counterparts while naturally providing uncertainty estimates for the reconstructed images. Furthermore, the estimated uncertainty maps showed a strong correlation with the true reconstruction errors and substantially outperformed deep learning-based uncertainty estimation. Another key advantage of the proposed framework is its ability to automatically estimate the model hyperparameters from the observed data, eliminating the need for manual tuning. Future work will investigate more expressive image priors, including deep generative priors, to further improve reconstruction quality while retaining the principled uncertainty quantification provided by Bayesian inference.

\bibliographystyle{IEEEtran}
\bibliography{mybib_filtered_noDublicates}

\newpage

 




\end{document}